\documentclass[lettersize,journal]{IEEEtran}
\usepackage{amsmath,amsfonts}
\usepackage{algorithmic}
\usepackage{array}
\usepackage[caption=false,font=normalsize,labelfont=sf,textfont=sf]{subfig}
\usepackage{textcomp}
\usepackage{stfloats}
\usepackage{url}
\usepackage{verbatim}
\usepackage{graphicx}
\usepackage{cite}
\usepackage{bbding}
\usepackage{hyperref}
\usepackage{color}
\usepackage{multirow}
\usepackage[ruled,vlined]{algorithm2e}
\def\BibTeX{{\rm B\kern-.05em{\sc i\kern-.025em b}\kern-.08em
		T\kern-.1667em\lower.7ex\hbox{E}\kern-.125emX}}
\usepackage{balance}
\begin{document}
	\title{Asynchronous Trajectory Matching-Based Multimodal Maritime Data Fusion for Vessel Traffic Surveillance in Inland Waterways}
	\author{Yu Guo,
		Ryan Wen Liu, \IEEEmembership{Member, IEEE},
		Jingxiang Qu,
		Yuxu Lu,\\
		Fenghua Zhu, \IEEEmembership{Senior Member, IEEE},
		and Yisheng Lv, \IEEEmembership{Senior Member, IEEE}
		
		\thanks{Yu Guo, Ryan Wen Liu, Jingxiang Qu, and Yuxu Lu are with the School of Navigation, Wuhan University of Technology, Wuhan 430063, China, and also with the National Engineering Research Center for Water Transport Safety, Wuhan, 430063, China (e-mail: \{yuguo, wenliu, qujx, yuxulu\}@whut.edu.cn).}
		\thanks{Fenghua Zhu and Yisheng Lv are with the State Key Laboratory of Management and Control for Complex Systems, Institute of Automation, Chinese Academy of Sciences, Beijing 100190, China (e-mail: \{fenghua.zhu, yisheng.lv\}@ia.ac.cn).}
		
	}
	
	
	\maketitle
	
	\begin{abstract}
		The automatic identification system (AIS) and video cameras have been widely exploited for vessel traffic surveillance in inland waterways. The AIS data could provide the vessel identity and dynamic information on vessel position and movements. In contrast, the video data could describe the visual appearances of moving vessels, but without knowing the information on identity, position and movements, etc. To further improve vessel traffic surveillance, it becomes necessary to fuse the AIS and video data to simultaneously capture the visual features, identity and dynamic information for the vessels of interest. However, traditional data fusion methods easily suffer from several potential limitations, e.g., asynchronous messages, missing data, random outliers, etc. In this work, we first extract the AIS- and video-based vessel trajectories, and then propose a deep learning-enabled asynchronous trajectory matching method (named DeepSORVF) to fuse the AIS-based vessel information with the corresponding visual targets. In addition, by combining the AIS- and video-based movement features, we also present a prior knowledge-driven anti-occlusion method to yield accurate and robust vessel tracking results under occlusion conditions. To validate the efficacy of our DeepSORVF, we have also constructed a new benchmark dataset (termed FVessel) for vessel detection, tracking, and data fusion. It consists of many videos and the corresponding AIS data collected in various weather conditions and locations. The experimental results have demonstrated that our method is capable of guaranteeing high-reliable data fusion and anti-occlusion vessel tracking.
	\end{abstract}
	
	\begin{IEEEkeywords}
		Inland waterways, vessel traffic surveillance, deep neural network, anti-occlusion tracking, data fusion.
	\end{IEEEkeywords}

	\section{Introduction}
	\IEEEPARstart{T}{he} autonomy and intelligence of the inland waterways surveillance system play a significant role in the development of inland waterborne transportation, which can effectively reduce the labor cost of the supervisory departments and ensure the safety of vessel navigation. To accomplish this objective, the vessel traffic service (VTS) system is capable of providing effective situational awareness by using the automatic identification system (AIS), radar, closed circuit television (CCTV) \cite{bloisi2016enhancing}, etc. To further increase the capability of situational awareness, many intelligent technologies for single sensor have been presented, e.g., AIS-based vessel trajectory prediction \cite{zhang2022vessel}, radar-based object detection \cite{kim2021bernoulli}, and video-based object detection \cite{feng2022rapid}. It is well known that each type of sensor has its own advantages and disadvantages under the same scenarios. As a consequence, numerous efforts have been devoted to simultaneously exploiting the multi-source data \cite{bloisi2016enhancing, chen2008tracking, man2016information, lu2021fusion, huang2021identity, liu2022intelligent} to promote the traffic situational awareness for maritime transportation systems. However, these fusion methods mainly just take into consideration the positional relationship of the same target at a certain moment. It thus becomes difficult to guarantee high-quality data fusion, especially for the existence of time delay, missing data, random outliers, etc. The same moving vessels essentially share similar navigation behaviors, which could be represented using the time-series data, e.g., spatio-temporal trajectories. To further improve the stability and accuracy of data fusion, we will first extract the vessel trajectories from the raw sensing data, and then propose a trajectory matching-based fusion method (termed DeepSORVF) in this work. 
	\subsection{Motivation and Contribution}
	Owing to the remote, intuitive, and real-time advantages of CCTV, terrestrial video surveillance systems have been widely used in inland waterborne transportation to improve the ability of traffic situational awareness and vessel abnormal behavior monitoring\cite{el2013target}. In particular, massive monitoring cameras can provide indispensable visual information for guaranteeing maritime safety. To fully use these visual features, many efforts have focused on the research of vessel detection and tracking to meet the requirement of intelligent supervision \cite{zhou2021deep, liu2021enhanced, chen2021visual}. However, these methods could only detect the moving vessel from the video images. It is intractable to achieve the important identity information (e.g., vessel name and size, etc.) and dynamic information (e.g., vessel speed and course, etc.).
	Other maritime awareness equipment, such as AIS and radar, could provide much richer attribute information about the vessel. In particular, the AIS data contains rich vessel identity and spatio-temporal information, which makes it play an essential role in analyzing vessel abnormal behaviors. The AIS data mainly contains the static and dynamic information, e.g., maritime mobile service identification (MMSI), vessel size, speed, course, position, etc. However, the AIS data essentially suffers from the inconsistency of time intervals, which limits its application in maritime intelligent transportation \cite{gao2021novel}. The radar has been widely used in near port supervision since it can provide the accurate distance and bearing of vessels. Unfortunately, some radar equipment is forbidden to be installed in populated regions to avoid the high-frequency electromagnetic radiation harming the health of people \cite{bloisi2016enhancing}. In the literature \cite{xiao2019traffic}, many methods have been proposed to robustly and accurately fuse the AIS and radar data. As long we simultaneously collect the AIS, radar, and video data, we can directly adopt the existing advanced methods to fuse the AIS and radar data, and then implement fusion with the video data. Intuitively, the fusion of AIS and video data seems more difficult because of the different coordinate systems, asynchronous data collection, different data structures, etc. Therefore, we tend to only fuse the AIS and video data to enhance the traffic situational awareness for intelligent surveillance in inland waterways. 
	In this work, we propose a deep learning-based simple online and real-time vessel data fusion method (termed DeepSORVF) for promoting inland waterways surveillance. The main contributions of this work are as follows:
	\begin{itemize}
		\item We build two simple yet efficient methods to respectively extract the AIS- and video-based vessel trajectories for data fusion. To avoid the interference of vessel occlusion on video-based trajectory extraction, we propose a prior knowledge-driven anti-occlusion tracking method.
		\item We design a novel asynchronous trajectory matching method to achieve the robust fusion of AIS and video data. The proposed method adopts an enhanced fast dynamic time warping algorithm for trajectory similarity measure and employs an AIS/video association method to decrease the computational cost and increase the stability.
		\item We construct a public benchmark dataset (termed FVessel) for vessel detection, tracking, and data fusion, which consists of many videos and the corresponding AIS data collected in various weather conditions and locations.
	\end{itemize}
	To our best knowledge, our DeepSORVF is the first trajectory matching-based computational method to fuse the AIS and video data for inland waterways surveillance. Meanwhile, we have verified the effectiveness and robustness of the proposed method on our newly-developed FVessel dataset.
	\subsection{Organization}
	The rest of this paper is organized as follows. Section \ref{sec:rw} briefly reviews the recent research on object detection, tracking, AIS and video data fusion. In Section \ref{sec:shipsort}, the proposed data fusion framework is described in detail. Section \ref{sec:exp} implements extensive comprehensive experiments to demonstrate the effectiveness of our method. Finally, Section \ref{con} summarizes the main contributions of this work.
	\section{Related Works}
	\label{sec:rw}
	This section mainly introduces the recent studies related to our work, i.e., multi-object detection and tracking, AIS and video data fusion.
	\subsection{Multi-Object Detection and Tracking}
	Multi-object detection and tracking methods are generally divided into two categories, namely traditional and deep learning methods. Due to the particularity of the research issue, this section mainly reviews the related works on vessel detection and tracking.
	\subsubsection{Traditional Methods}
	Background subtraction (BS) is a classic object detection method. Although many BS-based methods are proposed to detect conventional objects, these methods still achieve poor precision in vessel detection \cite{prasad2018object}. To improve the precision of the BS-based method, Hu \emph{et al.} \cite{hu2011robust} designed a robust foreground detection and background update method to effectively reduce the influence of waves. Bloisi \emph{et al.} \cite{bloisi2014background} proposed an independent multi-modal background subtraction (IMBS) algorithm. In particular, this algorithm models highly dynamic backgrounds (e.g., water) by creating a ``discretization'' of an unknown distribution. Furthermore, other types of vessel detection methods are proposed. For instance, Zhu \emph{et al.} \cite{zhu2010novel} designed a hierarchical complete-based vessel detection approach for spaceborne optical images. Zhang \emph{et al.} \cite{zhang2017ship} proposed a vessel detection algorithm using the discrete cosine transform (DCT)-based Gaussian mixture model (GMM) for efficient visual maritime surveillance on non-stationary surface platforms. Chen \emph{et al.} \cite{chen2019robust} achieved the vessel object tracking using multi-view learning and sparse representation. Although many techniques have been introduced to improve the performance of detectors, hand-designed features still produce poor robustness in vessel detection. Meanwhile, the high computational complexity of some methods will hinder their practical applications.
	\begin{figure*}[ht]
		\centering
		\includegraphics[width=0.98\linewidth]{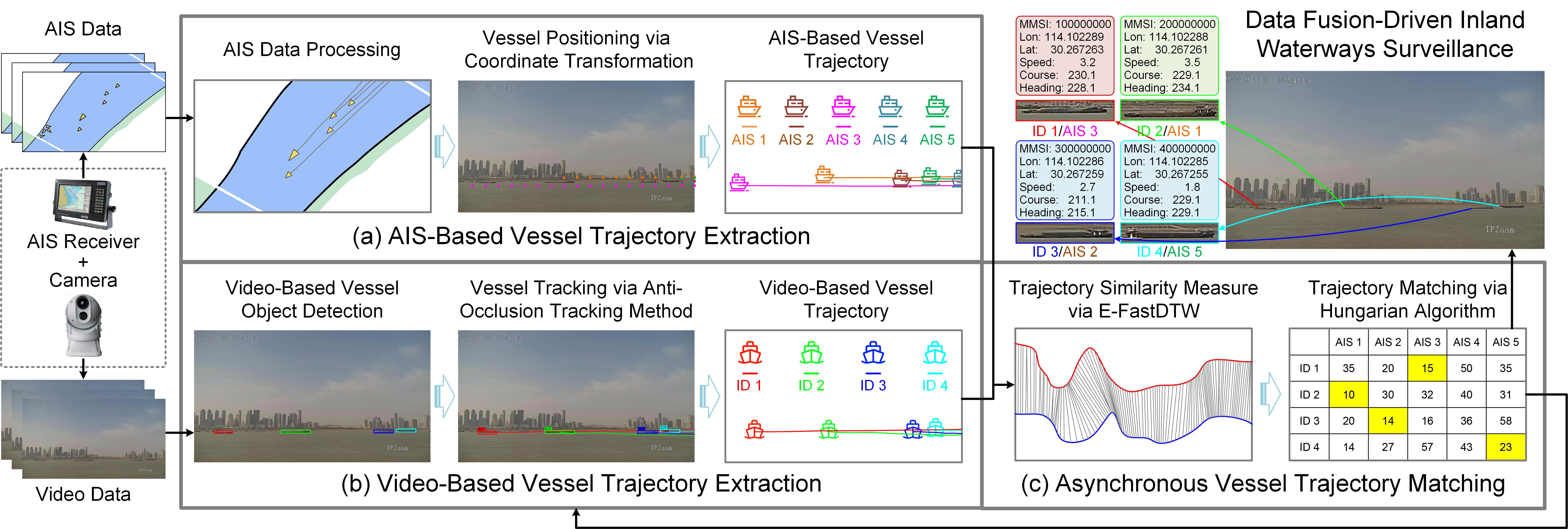}
		\caption{The architecture of the proposed deep learning-based simple online and real-time vessel data fusion method (termed DeepSORVF). The DeepSORVF consists of AIS-based vessel trajectory extraction, video-based vessel trajectory extraction, and asynchronous vessel trajectory matching.}
		\label{fig:flowchart}
	\end{figure*}
	\subsubsection{Deep Learning Methods}
	With the emergence and rapid development of graphics processing units (GPU), deep learning technology is widely used in the field of image processing. Many deep learning methods are proposed for object detection, e.g., region-based convolutional neural network (R-CNN) \cite{girshick2015region, ren2016faster}, single shot multibox detector (SSD) \cite{liu2016ssd, li2017fssd}, and you only look once network (YOLO) \cite{redmon2016you, redmon2017yolo9000, ge2021yolox, redmon2018yolov3}. Based on these object detection networks, many vessel detection methods are further researched. Shao \emph{et al.} \cite{shao2019saliency} proposed a YOLOv2-based saliency-aware network for vessel detection, which combined the salient features and coastline features to predict more accurate vessel positions. Liu \emph{et al.} \cite{liu2021enhanced} built an enhanced YOLOv3 network to promote vessel detection in video-based maritime surveillance. To reduce the impact of poor weather environments on vessel detection, this method constructed a data enhancement strategy to improve vessel detection precision in low-light, hazy, and rainy images. Furthermore, Chen \emph{et al.} \cite{chen2020deep} proposed a small vessel detection method based on an improved generative adversarial network (GAN) and a convolutional neural network (CNN). Feng \emph{et al.} \cite{feng2022rapid} proposed a ship detection method based on the multi-size gradient features and multi-branch support vector machine (SVM). Yang \emph{et al.} \cite{yang2021enhanced} applied the visual object tracking and semi-supervised object segmentation to the vessel tracking task, and proposed an enhanced SiamMask network. 
	\subsection{AIS and Video Data Fusion}
	In current literature, many AIS and video data fusion methods have been proposed. For instance, Chen \emph{et al.} \cite{chen2008tracking} proposed a single-vessel tracking method by combining AIS and video data. In particular, this method could make the camera focus on the vessel according to the position information provided by the AIS, and use the Kalman filter to ensure the smoothness of the tracking. However, the operator fails to accurately obtain the identities and attributes of each vessel when the field of view exists multiple vessels. Therefore, more researchers began to focus on the information fusion of multiple vessels. For instance, Man \emph{et al.} \cite{man2016information} fused the AIS and video data with the Kalman filter to obtain the optimal vessel trajectory. Bloisi \emph{et al.} \cite{bloisi2016enhancing} proposed an automated maritime surveillance system that replaces radar sensors with vision sensors, which can be deployed in densely populated regions. Lu \emph{et al.} \cite{lu2021fusion} proposed a vision and AIS fusion method, which estimated the distance and azimuth of the detected visual vessel from the camera and fused it with the position information in the AIS data. Huang \emph{et al.} \cite{huang2021identity} designed a novel multi-vessel tracking technology based on the improved single shot multi-box detector (SSD) \cite{liu2016ssd} and DeepSORT \cite{wojke2017simple} algorithm, and used a multi-modal data fusion algorithm to display the AIS information of visual targets. Recently, Liu \emph{et al.} \cite{liu2022intelligent} constructed an intelligent edge-enabled shipboard navigation system based on augmented reality, deep object detection, and multi-source data fusion technologies. This system can achieve stable vessel detection under various complex weather conditions and fuse the detected vessel targets with synchronized AIS information.
	\section{DeepSORVF: Deep Learning-based Simple Online and Real-time Vessel Data Fusion}
	\label{sec:shipsort}
	In this section, the details of our method will be introduced. Fig. \ref{fig:flowchart} displays the flowchart of our data fusion method, including AIS-based vessel trajectory extraction, video-based vessel trajectory extraction, and asynchronous vessel trajectory matching. For the AIS data, we perform data cleaning and delayed data prediction to obtain high-quality AIS data. To guarantee that the AIS and video data are in the same coordinate system, we use the pinhole model to project the AIS data to the pixel coordinate system. For the video data, we first use the YOLOX network to detect vessel targets. To avoid the impact of vessel occlusion on the video-based trajectory, a prior knowledge-driven anti-occlusion tracking method is then used for video-based vessel trajectory extraction. During trajectory matching, we adopt the enhanced fast dynamic time warping algorithm (E-FastDTW) to calculate the similarity between trajectories and combine the Hungarian algorithm to obtain the matching results. It is worth mentioning that the matching result will be input into the video-based vessel trajectory extraction task at the next moment as prior knowledge. Based on our matching results, the AIS information (including, MMSI, longitude, latitude, speed, course, heading, etc.) and the visual vessel can be easily fused to facilitate inland waterways surveillance.
	\begin{figure}[t]
		\centering
		\includegraphics[width=1\linewidth]{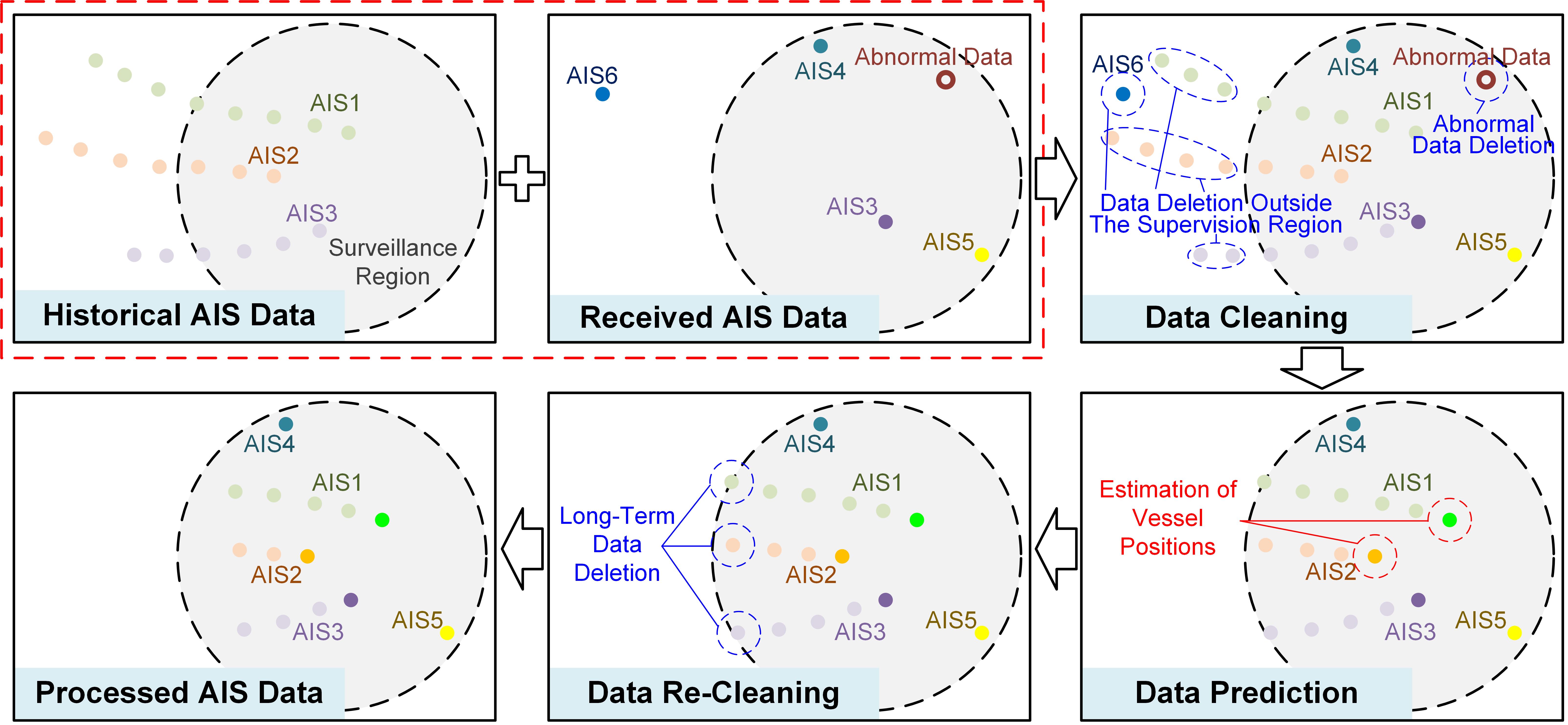}
		\caption{The flowchart of the AIS data processing, which consists of data cleaning, data prediction, and data re-cleaning.}
		\label{fig:AIS}
	\end{figure}
	\subsection{AIS-Based Vessel Trajectory Extraction}
	\label{se:AIS}
	The AIS is widely used in maritime services since it can provide the integrated and rich vessel information. However, due to the limitation of the AIS working principle, AIS data fails to be required in real-time. Meanwhile, some abnormal and redundant AIS information will affect the accuracy and robustness of AIS-based trajectory extraction. Furthermore, the matching of these two data becomes increasingly difficult since AIS- and video-based trajectories are, respectively, in the WGS-84 and pixel coordinate systems. Therefore, we tend to extract the AIS-based trajectory projected in the pixel coordinate system for data fusion. To achieve this goal, we first process the AIS data to generate high-quality AIS data. Subsequently, the AIS-based vessel trajectory will be obtained by the pinhole model.
	\subsubsection{AIS Data Processing}
	Fig. \ref{fig:AIS} displays our framework for processing AIS data. The historically processed AIS data and the AIS data received at the current moment are combined as the input. The input data is successively processed by data cleaning, data prediction, and data re-cleaning to obtain high-quality AIS data as output. The data cleaning process is used to delete the AIS data outside the supervision region and abnormal data including the missing and abnormalities of the latitude, longitude, heading, speed, and MMSI. The data prediction module can estimate the position of vessels that have not yet received AIS information. Let $v_{t-1}$ be the speed of the vessel at time $T_{t-1}$, the moving distance $D_{\Delta t}$ at the time interval $\Delta t = T_{t}-T_{t-1}$ can be expressed as $D_{\Delta t} = v_{t-1}*\Delta t$. According to the longitude $\lambda_{t-1}$, latitude $\phi_{t-1}$, and course $\theta_{t-1}$ at time $T_{t-1}$, and the moving distance at time interval $D_{\Delta t}$, the longitude $\lambda_{t}$ and latitude $\phi_{t}$ at time $T_{t}$ can be generated by the forward geodetic computations\footnote{We exploit the pyproj.Geod.fwd function to implement the forward geodetic computations.}.
	\begin{figure*}[ht]
		\centering
		\includegraphics[width=0.98\linewidth]{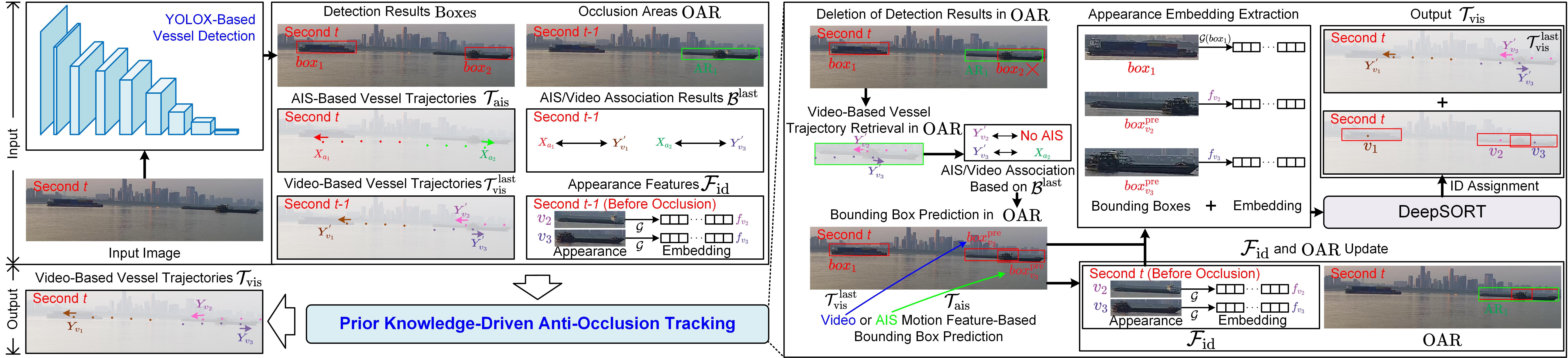}
		\caption{The flowchart of anti-occlusion tracking method for video-based vessel trajectory extraction. Note that $\mathcal{G}$ is the wide residual network-based appearance feature extractor. The extraction of AIS-based vessel trajectories $\mathcal{T}_{\mathrm{ais}}$ has been introduced in Section \ref{se:AIS}. The generation of Boxes, OAR, $\mathcal{T}^{\mathrm{last}}_{\mathrm{vis}}$, and $\mathcal{F}_{\mathrm{id}}$ will be mentioned in Section \ref{se:video}. The generation of AIS/video association results $\mathcal{B}^{\mathrm{last}}$ will be described in Section \ref{se:fusion}.}
		\label{fig:video}
	\end{figure*}
	\subsubsection{Vessel Positioning via Coordinate Transformation}
	To fuse the high-quality AIS information and visual object, it is necessary to unify the different source data into the same coordinate system. In this work, we tend to project the AIS information in the world coordinate system (WCS) to the pixel coordinate system (PCS). Before the coordinate transformation, we first perform a Mercator projection on the original position of the AIS information. Let $(U, V, W)$ be the real vessel position in the 3D WCS, its 2D projection coordinate $(x, y)$ in PCS can be obtained by
	\begin{equation}\label{eq:noisymodeadditive}
	\begin{bmatrix}	x\\ y\\ 1 \end{bmatrix} =  \frac{1}{Z} \mathcal{K}_{\mathrm{in}}\mathcal{K}_{\mathrm{ex}}\begin{bmatrix}	U\\ V\\ W\\ 1 \end{bmatrix},
	\end{equation}
	with $Z$ being the scale factor. Here, $\mathcal{K}_{\mathrm{in}}$ and $\mathcal{K}_{\mathrm{ex}}$ are the internal and external parameter matrices of the camera, respectively. In this work, since the camera is fixed, we set the extrinsic parameter matrix $\mathcal{K}_{\mathrm{ex}}$ as an identity matrix. In particular, we directly use the pinhole model to estimate $\mathcal{K}_{\mathrm{in}}$. Please refer to Ref. \cite{palmieri2013harbour} for more details on the internal parameter estimation. Finally, we sequentially save the AIS data with the same MMSI into the same list in time series to build a set of all AIS-based vessel trajectories $\mathcal{T}_{\mathrm{ais}} = \{X_{a_1},..., X_{a_i},..., X_{a_I}\}$ with $X_{a_i}$ and $I$ being the $i$-th AIS trajectory and the number of AIS-based vessel trajectories.
	\subsection{Video-Based Vessel Trajectory Extraction}
	\label{se:video}
	Although many methods have been proposed to achieve vessel detection and tracking \cite{feng2022rapid, prasad2019object, bovcon2021mods}, it is still intractable to extract high-quality video-based vessel trajectories for data fusion. In the actual application of video-based maritime surveillance, the inevitable occlusion between vessels occurs in the cross encounter, confrontation, and chasing situations. Generally, it becomes difficult to accurately and robustly detect these vessels under the occlusion condition. Meanwhile, the corresponding appearance will be seriously affected by other vessels. To improve the quality of extracted trajectories, we propose a prior knowledge-driven anti-occlusion tracking method, as shown in Fig. \ref{fig:video}.
	Specifically, we first adopt the YOLOX network to detect the visual vessel object and get a set of bounding boxes, i.e.,
	\begin{equation}\label{eq:Boxes}
	\mathrm{Boxes} = \{box_1,...,box_l,...,box_L\},
	\end{equation}
	where $box_l$ is the location of the $l$-th bounding box, $L$ denotes the number of bounding boxes. 
	\begin{algorithm}[t]
		\footnotesize
		\caption{Anti-Occlusion Vessel Tracking}
		\label{al:anti-occlusion}
		\LinesNumbered
		\KwIn {OAR: A set of all occlusion areas; 
			Boxes: A set of all bounding boxes detected by YOLOX network;
			$\mathcal{T}_{\mathrm{ais}}$: A set of all AIS-based vessel trajectories at the current moment;
			$\mathcal{T}^{\mathrm{last}}_{\mathrm{vis}}$: A set of all video-based vessel trajectories at the previous moment;
			$\mathcal{B}^{\mathrm{last}}$: A set of all AIS/video association results at the previous moment;
			$\mathcal{F}_{\mathrm{id}}$: A set of all occluded vessel appearance features before the occlusion;}
		\KwOut {$\mathcal{T}_{\mathrm{vis}}$: A set of all video-based vessel trajectories at the current moment;} 
		\textbf{Initialization:} 
		Boxes$^{\mathrm{pre}}$: A empty set to save all predict bounding boxes of the occluded vessels;
		$\mathcal{G}$: A wide residual network-based appearance feature extractor;
		$\mathcal{I}$: A empty set to save all DeepSORT input data\;
		\tcp{Step 1. Bounding box removal in the occlusion areas.}
		\For{$box_l$ in $\mathrm{Boxes}$}{
			\For{$\mathrm{AR}$ in $\mathrm{OAR}$}{
				\uIf{$box_l$ locates in $\mathrm{AR}$}{
					Remove the $box_l$ from $\mathrm{Boxes}$\;
					\textbf{break}\;}
		}}
		\tcp{Step 2. Occluded bounding box prediction.}
		\For{$Y^{'}_{v_j}$ in $\mathcal{T}^{\mathrm{last}}_{\mathrm{vis}}$}{
			\For{$\mathrm{AR}$ in $\mathrm{OAR}$}{
				\uIf{The center point of the bounding box at the previous moment in $Y^{'}_{v_j}$ locates in $\mathrm{AR}$}{Search the matched $[:,v_{j}]$ from $\mathcal{B}^{\mathrm{last}}$\;
					\uIf{exist $[a_{i},v_{j}]$}{
						Predict the bounding box $box^{\mathrm{pre}}_{v_{j}}$ by AIS-based vessel trajectory $X_{a_i}$ in $\mathcal{T}_{\mathrm{ais}}$\;}
					\Else{Predict the bounding box $box^{\mathrm{pre}}_{v_{j}}$ by video-based vessel trajectory$Y^{'}_{v_j}$\;
					}
					Add the $box^{\mathrm{pre}}_{v_{j}}$ to $\mathrm{Boxes}^{\mathrm{pre}}$\;
					\textbf{break}\;}
				
			}
		}
		Update $\mathcal{F}_{\mathrm{id}}$ and $\mathrm{OAR}$\;
		\tcp{Step 3. Anti-occlusion DeepSORT.}
		\For{$box_l$ in $\mathrm{Boxes}$}{
			Add $[box_l, \mathcal{G}(box_l)]$ to $\mathcal{I}$\;
		}
		\For{$box^{\mathrm{pre}}_{v_{j}}$ in $\mathrm{Boxes}^{\mathrm{pre}}$}{
			\For{$f_{v_b}$ in $\mathcal{F}_{\mathrm{id}}$}{
				\uIf{$v_j = v_b$}{
					Add $[box^{\mathrm{pre}}_{v_{j}}, f_{v_b}]$ to $\mathcal{I}$\;
					\textbf{break}\;}
		}}
		Run DeepSORT with $\mathcal{I}$\;
		Add the results of DeepSORT to $\mathcal{T}^{\mathrm{last}}_{\mathrm{vis}}$ for generating the video-based vessel trajectory at the current moment $\mathcal{T}_{\mathrm{vis}}$\;
	\end{algorithm}
	Before the tracking, the results of the previous moment are input as the prior knowledge. Firstly, a set of the occlusion areas OAR will be used, which depends on the ratio of the occlusion area to the bounding boxes. The judgment metric of the occlusion area can be expressed as follows
	\begin{equation}\label{eq:OAR}
	\frac{S_o}{\min(S_1, ..., S_r,... ,S_R)} > \omega,
	\end{equation}
	where $S_o$ is the area of the occluded part, $S_r$ is the area of the $r$-th occluded bounding box, $R$ is the number of occluded bounding boxes, $\omega$ represents the anti-occlusion threshold. When the ratio exceeds $\omega$, we will store the location of the smallest rectangle box AR which can contain all occluded bounding boxes into the OAR. Meanwhile, the AIS-based vessel trajectories at the current moment $\mathcal{T}_{\mathrm{ais}}$ and the video-based vessel trajectories at the previous moment $\mathcal{T}^{\mathrm{last}}_{\mathrm{vis}}$ are also used as prior information, which can be given by
	\begin{equation}\label{eq:aisvideo}
	\begin{cases}
	&\mathcal{T}_{\mathrm{ais}} = \{X_{a_1},...,X_{a_i},...,X_{a_I}\},\\
	&\mathcal{T}^{\mathrm{last}}_{\mathrm{vis}} = \{Y^{'}_{v_1},...,Y^{'}_{v_j},...,Y^{'}_{v_J}\},\\
	\end{cases}
	\end{equation}
	where $X_{a_i}$ and $Y^{'}_{v_j}$ represent the trajectory series of the $i$-th AIS target and the $j$-th visual target, respectively, $I$ and $J$ are the numbers of AIS- and video-based vessel trajectories, respectively. Besides, we consider the AIS/video association results $\mathcal{B}^{\mathrm{last}}$ at the previous moment and the vessel appearance embedding $\mathcal{F}^{\mathrm{last}}_{\mathrm{id}}$ before the occlusion, which can be given by
	\begin{equation}\label{eq:MF}
	\begin{cases}
	&\mathcal{B}^{\mathrm{last}} = \{...,[a_i,v_j],...\},\\
	&\mathcal{F}_{\mathrm{id}} = \{...,f_{v_b},...\},\\
	\end{cases}
	\end{equation}
	where $[a_i,v_j]$ means that the $i$-th AIS target $a_i$ and $j$-th visual target $v_j$ are successfully associated, $f_{v_b}$ is the appearance embedding of the $b$-th visual target before occlusion\footnote{The vessel appearance embedding is extracted by the wide residual network \cite{zagoruyko2016wide} in the DeepSORT.}. We will detailedly describe the generation of AIS/video association results $\mathcal{B}^{\mathrm{last}}$ in Section \ref{se:fusion}.
	For the anti-occlusion tracking, the detection results located in the occlusion area OAR are removed to avoid the mis-detection caused by the vessel overlapping. Based on the fusion results at the previous moment, the corresponding AIS information of the occluded visual vessel is available. Therefore, the location of the occluded vessel's bounding box at the current moment $box^{\mathrm{pre}}$ can be estimated by
	\begin{equation}\label{eq:position_pre}
	box^{\mathrm{pre}} = \begin{bmatrix} x\mathrm{_{tl}^{pre}}\\ y\mathrm{_{tl}^{pre}}\\ x\mathrm{_{br}^{pre}} \\ y\mathrm{_{br}^{pre}}\end{bmatrix} =  \begin{bmatrix}	\Delta x_{\mathrm{ais}}\\ \Delta y_{\mathrm{ais}}\\ \Delta x_{\mathrm{ais}}\\ \Delta y_{\mathrm{ais}} \end{bmatrix} + \begin{bmatrix}	x\mathrm{_{tl}^{last}}\\ y\mathrm{_{tl}^{last}}\\ x\mathrm{_{br}^{last}}\\ y\mathrm{_{br}^{last}}\end{bmatrix},
	\end{equation}
	where $(x\mathrm{_{tl}^{last}}, y\mathrm{_{tl}^{last}})$ and $(x\mathrm{_{br}^{last}}, y\mathrm{_{br}^{last}})$ are the pixel indexes of the top-left and bottom-right points of the previous bounding box, respectively, $\Delta x_{\mathrm{ais}}$ and $\Delta y_{\mathrm{ais}}$ are the horizontal and vertical motion speeds, which are equal to the displacement of the AIS information between the current and previous moments.
	For the occluded vessels without the corresponding AIS information, the bounding box will be predicted via the visual motion features. The prediction result can also be given by variants based on Eq. (\ref{eq:position_pre}). The horizontal and vertical motion speeds ($\Delta x_{\mathrm{ais}}$, $\Delta y_{\mathrm{ais}}$) are replaced with the visual trajectory-based horizontal and vertical motion speeds ($\Delta x_{\mathrm{vis}}$, $\Delta y_{\mathrm{vis}}$), which can be calculated by
	\begin{equation}\label{eq:visual_kin}
	\begin{cases}
	&\Delta x_{\mathrm{vis}} = \dfrac{x_{t-1} -  x_{t-\delta}}{\delta},\\
	&\Delta y_{\mathrm{vis}} = \dfrac{y_{t-1} -  y_{t-\delta}}{\delta},\\
	\end{cases}
	\end{equation}
	where $(x_{t-1}, y_{t-1})$ and $(x_{t-\delta}, y_{t-\delta})$ denote the points of the video-based vessel trajectory at the previous moment and the previous $\delta$ moment, respectively. 
	After prediction, we will then update the OAR and $\mathcal{F}_{\mathrm{id}}$. Based on the predicted detection box position, the occlusion area list OAR will be updated via Eq. (\ref{eq:OAR}) for the anti-occlusion tracking at the next moment. For the update of $\mathcal{F}_{\mathrm{id}}$, we first set up the occluded visual target as $v_j$. If $f_{v_j}$ exists in the original $\mathcal{F}_{\mathrm{id}}$, we directly store $f_{v_j}$ into the new $\mathcal{F}_{\mathrm{id}}$; otherwise, the appearance embedding of $v_j$ at the previous moment will be stored in the new $\mathcal{F}_{\mathrm{id}}$. Then, we employ a wide residual network $\mathcal{G}$ to extract the vessel appearance embedding in normal bounding boxes, and assign the vessel appearance embedding before occlusion in the $\mathcal{F}_{\mathrm{id}}$ to occluded bounding boxes. Finally, the bounding boxes and the corresponding vessel appearance embedding are jointly input into the DeepSORT for generating the video-based vessel trajectories at the current moment $\mathcal{T}_{\mathrm{vis}}$.
	It is worth mentioning that two metrics in DeepSORT can solve the ID assignment issue. Firstly, the Mahalanobis distances between the predicted Kalman states and the newly arrived locations are calculated as the location similarity metrics. Moreover, the cosine distances between the appearance embedding are calculated as the appearance similarity metrics. In our method, the appearance features of the occluded vessels are kept consistent with the latest extractions before the occlusion. Therefore, as long as the predicted bounding box is close to the prediction of Kalman filters, the ID of occluded vessels will not be assigned incorrectly. The pseudo code of the proposed anti-occlusion tracking method is shown in Algorithm \ref{al:anti-occlusion}.
	\subsection{Asynchronous Vessel Trajectory Matching}
	\label{se:fusion}
	In this section, we propose a simple yet effective trajectory matching method to fuse the AIS- and video-based asynchronous vessel trajectories. Firstly, we adopt an enhanced fast dynamic time warping (E-FastDTW) algorithm considering the direction to calculate the similarity of AIS- and video-based vessel trajectories. Based on the similarity measure result, the Hungarian algorithm is employed to generate the optimal matching result. To improve the stability and robustness of data fusion and reduce the computational cost, we employ an AIS/video association mechanism. When the number of successful pairings of two trajectories exceeds a pre-determined threshold, the AIS- and video-based vessel trajectories will be associated directly without similarity evaluation.
	\subsubsection{Trajectory Similarity Measure via E-FastDTW}
	For trajectory-based data fusion, it is an important prerequisite to determine the similarities between the AIS- and video-based vessel trajectories. The Euclidean distance is a simple but effective similarity calculation method. However, it requires that the two trajectories to be matched have the same length. Meanwhile, the Euclidean distance considers that two similar trajectories with only a slight shift in the time axis are significantly different. Therefore, dynamic time warping (DTW) has been proposed for ignoring this shift \cite{muller2007dynamic}. Suppose we have two trajectories $X$ and $Y$ of length $P$ and $Q$ respectively, represented as
	\begin{equation}\label{eq:series}
	\begin{cases}
	&X = m_1, m_2,...,m_p,...,m_P,\\
	&Y = n_1, n_2,...,n_q,...,n_Q.\\
	\end{cases}
	\end{equation}
	\begin{algorithm}[t]
		\footnotesize
		\caption{Asynchronous Trajectory Matching}
		\label{al:matching}
		\LinesNumbered
		\KwIn {$\mathcal{T}_{\mathrm{ais}}$: A set of all AIS-based vessel trajectories; $\mathcal{T}_{\mathrm{vis}}$: A set of all video-based vessel trajectories; $\mathcal{M}^{\mathrm{last}}$: A set of all AIS/video numbers of matches at the previous moment; $\mathcal{B}^{\mathrm{last}}$: A set of all AIS/video association result at the previous moment;}
		\KwOut {$\mathcal{M}$: A set of all AIS/video numbers of matches at the current moment; $\mathcal{B}$: A set of all AIS/video association result at the current moment;}
		\textbf{Initialization:} $d(i, j)$: The Euclidean distance between the last trajectory points of $X_{a_i}$ and $Y_{v_j}$; $M_s$: An empty trajectory similarity matrix; $\mathcal{O}_{\mathrm{res}}$: An empty set to save the matching results; $D_{\mathrm{max}}$: The maximum matching distance; $Mat_{\mathrm{min}}$: The minimum number of matches; $T_{\mathrm{max}}$: The maximum time threshold; $S$: The E-FastDTW trajectory similarity measurement operator\;
		
		\tcp{Step 1. Trajectory similarity measure.}
		\For{$X_{a_i}$ in $\mathcal{T}_{\mathrm{ais}}$}{
			\For{$Y_{v_j}$ in $\mathcal{T}_{\mathrm{vis}}$}{ 
				\uIf{$d(i, j)>D_{\mathrm{max}}$}{  
					$M_{s}(i, j) = + \infty$ \;}
				\uElseIf{$[a_i,:]$ or $[:,v_j]$ in $\mathcal{B}^{\mathrm{last}}$}{
					\uIf{$[a_i, v_j]$ in $\mathcal{B}^{\mathrm{last}}$}{
						$M_{s}(i, j) = - \infty$ \;}
					\Else{
						$M_{s}(i, j) = + \infty$ \;}}
				\Else{
					$M_{s}(i, j) = S(X_{a_i}, Y_{v_j})$ by Eq. (\ref{eq:dtw})\;}
		}}
		\tcp{Step 2. Matching result generation.}
		Using the Hungarian algorithm to calculate $M_{s}$ for obtaining the matching result $\mathcal{O}_{\mathrm{res}}=\{...,[a_{i}, v_{j}],...\}$\; 
		\For{$[a_i, v_j]$ in $\mathcal{O}_{\mathrm{res}}$}{
			\uIf{$z_{a_i, v_j}$ in $\mathcal{M}^{\mathrm{last}}$}{
				Add $z_{a_i, v_j} = z_{a_i, v_j}++$ to $\mathcal{M}$\;}
			\Else{
				Add $z_{a_i, v_j}=1$ to $\mathcal{M}$ \;}
		}
		\For{$z_{a_i, v_j}$ in $\mathcal{M}^\mathrm{last}$}{
			\uIf{$[a_i,v_j]$ not in $\mathcal{O}_{\mathrm{res}}$ and the time interval between the last matching moment of $[a_i, v_j]$ and the current moment $<T_{\mathrm{max}}$}{
				Add $z_{a_i, v_j}$ to $\mathcal{M}$\;}
			
		}
		\tcp{Step 3. Association result generation.}
		\For{$z_{a_i, v_j}$ in $\mathcal{M}$}{
			\uIf{$z_{a_i, v_j}>Mat_{\mathrm{min}}$}{
				Add $[a_i, v_j]$ to $\mathcal{B}$\;
		}}
	\end{algorithm}
	Based on the two trajectories, the DTW constructs a $P \times Q$ alignment matrix $d$ where $d(p, q)$ is the Euclidean distance between the points $m_p$ and $n_q$. Then, a warp path $W$ is defined to construct the mapping between $X$ and $Y$, which can be written by
	\begin{equation}\label{eq:warppath}
	W = w_1,w_2,...,w_c,...,w_C,
	\end{equation}
	with $C$ being the length of $W$, and $\max\{P, Q\} \leq C < P+Q$. In particular, the warp path $W$ has three restrictions. For the sake of better understanding, we define the $(c-1)$-th and the $c$-th elements of $W$ as $w_{c-1}=(p^{'}, q^{'})$ and $w_c=(p, q)$. These three constraints for warp path can be defined as follows:
	\begin{itemize}
		\item \emph{Restriction 1:} The $1$-st and the $C$-th elements of $W$ are $w_1=(1, 1)$ and $w_{C}=(P,Q)$, respectively.
		\item \emph{Restriction 2:} The adjacent elements of the warp path $W$ can only contain the adjacent coordinate points, including the diagonal adjacent. Therefore, the $w_{c-1}$ can only be one of $\left\{ (p-1, q), ~(p, q-1), ~(p-1, q-1) \right\}$.
		\item \emph{Restriction 3:} The elements of the warp path $W$ are monotonically increasing in time, i.e., $p^{'} \leq p$ and $q^{'} \leq q$. 
	\end{itemize}
	Under the premise of satisfying the above three constraints, DTW only focuses on the path with the minimum cumulative distance of alignment matrix elements corresponding to all points \cite{muller2007dynamic}. Meanwhile, the included angle $\varphi$ between the starting and ending points of $X$ and $Y$ is also considered. Finally, the similarity value $S(X, Y)$ between $X$ and $Y$ calculated by our proposed E-FastDTW can be written as follows
	\begin{equation}\label{eq:dtw}
	S(X, Y) = Dis(W) \cdot e^{\varphi} = \min\{\sum_{c=1}^{C}d(w_{cp},w_{cq})\} \cdot e^{\varphi},
	\end{equation}
	where $d(w_{cp},w_{cq})$ is the Euclidean distance between two data points corresponding to the $c$-th element in the warp path $W$, $Dis(W)$ denotes the sum of all $d(w_{cp},w_{cq})$ in the warp path $W$. To find the desired unique warp path, the DTW adopts the dynamic programming strategy. The cumulative distance $\mathcal{D}(p, q)$ between $m_p$ and $n_q$ is the sum of the minimum cumulative distance of three previous possible warp path elements and the Euclidean distance $d(p, q)$ between the points $m_p$ and $n_q$, which can be mathematically written as
	\begin{equation}\label{eq:dynamic}
	\begin{aligned}
	\mathcal{D}(p, q) = d(p, q)& + \min\{\mathcal{D}(p-1, q), \\
	&\mathcal{D}(p, q-1), \mathcal{D}(p-1, q-1)\}.
	\end{aligned}
	\end{equation}
	Furthermore, we also adopt the multi-level approach used in the FastDTW to speed up the time series similarity search and reduce the computational complexity. Please refer to  Ref. \cite{salvador2007toward} for more details on the multi-level approach.
	\subsubsection{Trajectory Matching}
	In this work, we propose a novel matching method with higher precision and less computation. In particular, we will match and associate the AIS-based vessel trajectories $\mathcal{T}_{\mathrm{ais}}$ mentioned in Section \ref{se:AIS}, and the video-based vessel trajectories $\mathcal{T}_{\mathrm{vis}}$ mentioned in Section \ref{se:video}, which can be defined as follows
	\begin{equation}\label{eq:trajectory}
	\begin{cases}
	&\mathcal{T}_{\mathrm{ais}} = \{X_{a_1},..., X_{a_i},..., X_{a_I}\},\\
	&\mathcal{T}_{\mathrm{vis}} = \{Y_{v_1},..., Y_{v_j},..., Y_{v_J}\},\\
	\end{cases}
	\end{equation}
	where $X_{a_i}$ and $Y_{v_j}$ represent the trajectories of the $i$-th AIS target $a_i$ and the $j$-th visual target $v_j$, respectively, $I$ and $J$ are the numbers of AIS- and video-based vessel trajectories, respectively. Furthermore, the numbers of AIS/video matches $\mathcal{M}^{\mathrm{last}}$ and association results $\mathcal{B}^{\mathrm{last}}$ at the previous moment are also considered as input, i.e.,
	\begin{equation}\label{eq:fusion}
	\begin{cases}
	&\mathcal{M}^{\mathrm{last}} = \{..., z_{a_i, v_j},...\},\\
	&\mathcal{B}^{\mathrm{last}} = \{..., [a_i, v_j],...\},\\
	\end{cases}
	\end{equation}
	where $z_{a_i, v_j}$ is the number of successful matches of $X_{a_i}$ and $Y_{v_j}$, $[a_i,v_j]$ means that $a_i$ and $v_j$ have been associated together. In the similarity measure, it is obviously time-consuming and intractable to adopt the E-FastDTW for calculating the similarity between all trajectories at each moment. Inspired by the DeepSORT algorithm, we propose a trajectory association mechanism to solve these issues. In particular, if two trajectories have been recorded in the $\mathcal{B}^{last}$, the two trajectories are directly matched by default without similarity measurement with other trajectories. Subsequently, we perform the similarity measure between all trajectories and construct a similarity matrix $M_s$ of size $I \times J$, where $M_s(i,j)$ represents the similarity value of $X_{a_i}$ and $Y_{v_j}$. In particular, when the Euclidean distance between the last trajectory points of $X_{a_i}$ and $Y_{v_j}$ exceeds the maximum matching distance $D_{\mathrm{max}}$, we consider the two trajectories to be completely different and set $M_s(i, j) = + \infty$. When the binding trajectory pair $[a_i,v_j]$ exists in the $\mathcal{B}^{\mathrm{last}}$, we set $M_s(i, j)= - \infty$ and set the values of other horizontal and vertical positions to positive infinity. For other ordinary trajectory pairs that do not satisfy the above conditions, we employ Eq. (\ref{eq:dtw}) (i.e., E-FastDTW) to calculate the trajectory similarity. After obtaining the similarity matrix $M_s$, we adopt the Hungarian optimization algorithm to find the optimal matching result $\mathcal{O}_{\mathrm{res}}$, which contains the matching trajectory pair information, i.e., 
	\begin{equation}\label{eq:result}
	\mathcal{O}_{\mathrm{res}}=\{...,[a_{i}, v_{j}],...\},
	\end{equation}
	where $[a_{i}, v_{j}]$ means that $a_i$ and $v_j$ are matched together. Then, we will generate the AIS/video matching results $\mathcal{M}$ and association results $\mathcal{B}$ at the current moment. More specifically, we iterate through all matching trajectory pairs in the $\mathcal{O}_{\mathrm{res}}$. If the number of matching times $z_{a_i, v_j}$ of trajectory pair $[a_i, v_j]$ in the $\mathcal{O}_{\mathrm{res}}$ exists in the $\mathcal{M}^\mathrm{last}$, we will store $(z_{a_i, v_j}+1)$ to $\mathcal{M}$; otherwise, 1 will be stored to $\mathcal{M}$. In addition, we save the number of matching times $z_{a_i,v_j}$ for some trajectory pairs directly from $\mathcal{M}^\mathrm{last}$ into $\mathcal{M}$. These $z_{a_i,v_j}$ need to satisfy two conditions, which can be defined as follows: 
	\begin{itemize}
		\item $z_{a_i,v_j}$ must exist in $\mathcal{M}^\mathrm{last}$ but $[a_i,v_j]$ is not in $\mathcal{O}_{\mathrm{res}}$.
		\item The time interval between the last matching moment and the current moment is less than $T_{\mathrm{max}}$.
	\end{itemize}
	For the generation of the AIS/video association result, we set a minimum number of matches $Mat_{\mathrm{min}}$ as a threshold to ensure that the association information is accurate. When $z_{a_i, v_j}$ in the $\mathcal{M}$ is greater than $Mat_{\mathrm{min}}$, we will store $[a_i, v_j]$ into $\mathcal{B}$. The pseudo code of the proposed trajectory matching method is shown in Algorithm \ref{al:matching}. 
	\setlength{\tabcolsep}{6pt}
	\begin{table}[t]
		\scriptsize
		\centering
		\caption{Details of the FVessel dataset. The ``TOO'', ``NOV'', and ``NOA'' are the times of occlusions, the total number of vessels, and the number of vessels with AIS, respectively.}
		\begin{tabular}{c|cccccc}
			\hline
			Video    & Video Length & Type      & Weather   & TOO & NOV & NOA \\ \hline \hline
			video-01 & 10m07s       & Bridge    & Low-light & 2   & 5   & 4   \\
			video-02 & 19m52s       & Bridge    & Sunny     & 6   & 7   & 6   \\
			video-03 & 19m14s       & Riverside & Sunny     & 6   & 5   & 5   \\
			video-04 & 06m10s       & Riverside & Sunny     & 0   & 1   & 1   \\
			video-05 & 15m01s       & Riverside & Sunny     & 2   & 5   & 5   \\
			video-06 & 12m49s       & Riverside & Sunny     & 2   & 4   & 4   \\
			video-07 & 03m38s       & Riverside & Sunny     & 1   & 2   & 2   \\
			video-08 & 16m05s       & Riverside & Sunny     & 3   & 6   & 5   \\
			video-09 & 05m25s       & Riverside & Sunny     & 0   & 1   & 1   \\
			video-10 & 11m17s       & Bridge    & Cloudy    & 2   & 3   & 1   \\
			video-11 & 05m18s       & Riverside & Sunny     & 1   & 3   & 3   \\
			video-12 & 07m19s       & Riverside & Sunny     & 1   & 4   & 4   \\
			video-13 & 12m58s       & Riverside & Sunny     & 5   & 6   & 6   \\
			video-14 & 03m58s       & Riverside & Sunny     & 3   & 4   & 4   \\
			video-15 & 10m46s       & Riverside & Sunny     & 0   & 4   & 4   \\
			video-16 & 05m05s       & Riverside & Sunny     & 0   & 1   & 1   \\
			video-17 & 08m08s       & Riverside & Sunny     & 1   & 2   & 2   \\
			video-18 & 23m57s       & Riverside & Sunny     & 10  & 10  & 6   \\
			video-19 & 11m28s       & Riverside & Low-light & 0   & 2   & 2   \\
			video-20 & 14m10s       & Riverside & Low-light & 0   & 3   & 3   \\
			video-21 & 24m01s       & Riverside & Low-light & 4   & 7   & 6   \\
			video-22 & 02m40s       & Riverside & Low-light & 0   & 2   & 1   \\
			video-23 & 19m24s       & Riverside & Sunny     & 2   & 4   & 4   \\
			video-24 & 08m39s       & Riverside & Sunny     & 2   & 3   & 3   \\
			video-25 & 24m05s       & Riverside & Sunny     & 4   & 8   & 8   \\
			video-26 & 07m26s       & Riverside & Sunny     & 0   & 5   & 5   \\ \hline
		\end{tabular}\label{table:FVessel}
	\end{table}
	\subsection{Implementation Details}
	This section mainly introduces the detailed settings of the proposed data fusion method. In particular, our method is implemented on the python 3.7 platform. All experiments and tests are conducted on a PC with Intel Core i5-10600KF CPU @ 4.10GHz and Nvidia RTX A4000 GPU. To meet the requirement of real-time processing while ensuring the accurate fusion, our method only executes one processing per second. For the AIS-based vessel trajectory extraction, we delete the data more than two nautical miles from the camera and set the maximum storage time to two minutes. For the vessel detection task, we collect 20k images containing vessel objects as the training dataset. In training, we set the epoch to 100 and employ the Adam algorithm as the optimizer. The initial learning rates for the first 50 and last 50 epochs are $10^{-3}$ and $10^{-4}$, respectively. For the video-based vessel trajectory extraction, we set the occlusion area threshold $\omega = 0$ and the time span of visual motion feature extraction $\delta = 5s$. For the AIS and video data fusion, we set the maximum matching distance $D_{\mathrm{max}}$ as the half of the horizontal size of the image, the minimum number of matching times $Mat_{\mathrm{min}} = 15$, and the maximum time threshold $T_{\mathrm{max}}=15s$.
	\section{Experimental Results and Discussion}
	\label{sec:exp}
	In this section, we conduct massive experiments on vessel detection, vessel tracking, and data fusion to quantitatively evaluate the performance of our proposed method. The running time analysis is also carried out to verify its practicality. 
	\subsection{Benchmark Dataset}
	\begin{figure}[t]
		\centering
		\includegraphics[width=1\linewidth]{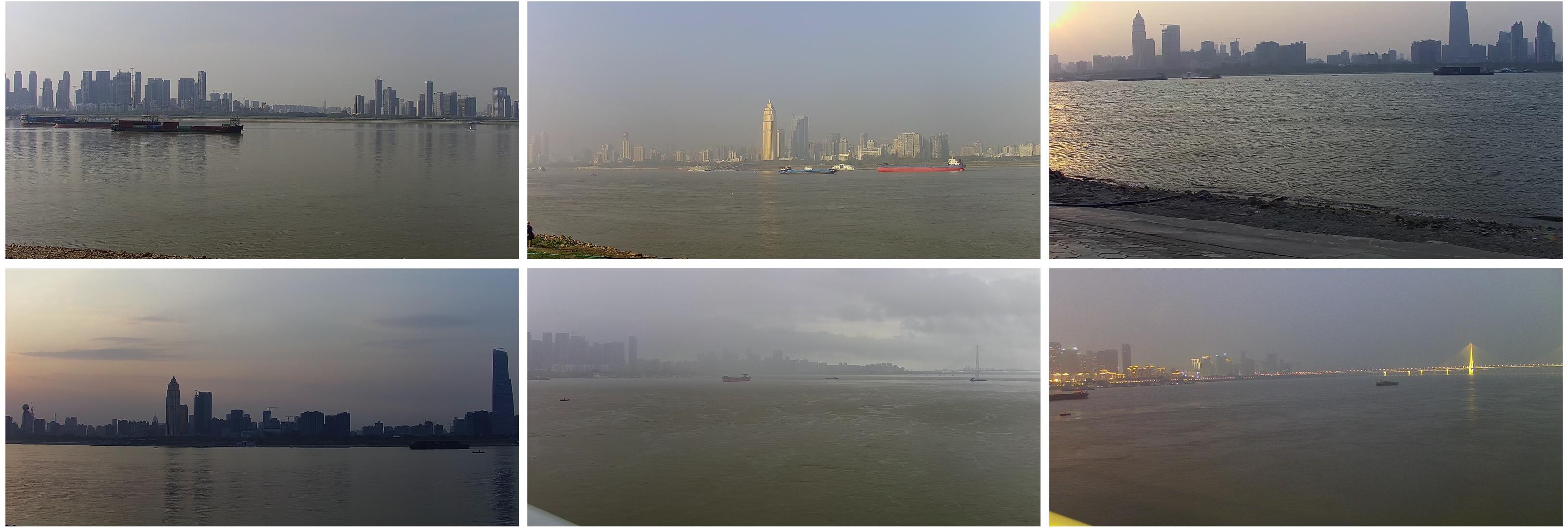}
		\caption{Some samples of the FVessel dataset, which contains massive images and videos captured on the bridge region and riverside under sunny, cloudy, and low-light conditions.}
		\label{fig:FVessel}
	\end{figure}
	\setlength{\tabcolsep}{6pt}
	\begin{table}[t]
		\scriptsize
		\centering
		\caption{Details of the dataset used in the vessel detection, vessel tracking, and data fusion experiments. The ``TOO'', ``NOV'', and ``NOA'' are the times of occlusions, the total number of vessels, and the number of vessels with AIS, respectively.}
		\begin{tabular}{c|cccccc}
			\hline
			Video   & Video Length & Type      & Weather & TOO & NOV & NOA \\ \hline \hline
			clip-01 & 01m51s       & Bridge    & Cloudy  & 1   & 2   & 2   \\
			clip-02 & 01m36s       & Riverside & Sunny   & 1   & 2   & 2   \\
			clip-03 & 03m42s       & Riverside & Low-light    & 1   & 2   & 2   \\
			clip-04 & 02m42s       & Riverside & Low-light    & 2   & 3   & 3   \\
			clip-05 & 03m05s       & Riverside & Sunny   & 2   & 3   & 3   \\
			clip-06 & 02m38s       & Riverside & Sunny   & 2   & 3   & 3   \\
			clip-07 & 11m10s       & Riverside & Sunny   & 2   & 4   & 4   \\
			clip-08 & 05m07s       & Riverside & Sunny   & 1   & 2   & 2   \\
			clip-09 & 08m39s       & Riverside & Sunny   & 2   & 3   & 3   \\
			clip-10 & 02m46s       & Riverside & Sunny   & 2   & 4   & 4   \\ \hline
		\end{tabular}\label{table:dataset}
	\end{table}
	In this section, we construct a benchmark dataset \footnote{We have released some matched pairs of AIS and video data, available at \url{https://github.com/gy65896/FVessel}. The complete dataset and original source code will be released if this work is finally accepted for publication.} for vessel detection, tracking, and data fusion (named FVessel) containing 26 videos and the corresponding AIS data captured by the HIKVISION DS-2DC4423IW-D dome camera and Saiyang AIS9000-08 Class-B AIS receiver on the Wuhan Segment of the Yangtze River \footnote{The vessel is uniquely identified by a particular MMSI in raw AIS data. To protect the privacy, the MMSI for each vessel has been replaced with a random number in our dataset.}. As shown in Fig. \ref{fig:FVessel}, these videos were captured under many locations (e.g., bridge region and riverside) and various weather conditions (e.g., sunny, cloudy, and low-light). Table \ref{table:FVessel} displays more details about the FVessel dataset, including the video length, collection location, weather condition, the times of occlusions, the total number of vessels, and the number of vessels with AIS. To verify the superiority of the proposed module, we intercept ten clips existing the vessel occlusion from the FVessel dataset for comparison experiments on vessel detection, tracking, and data fusion. More detailed information on the test dataset can be found in Table \ref{table:dataset}.
	\setlength{\tabcolsep}{6pt}
	\begin{table}[t]
		\scriptsize
		\centering
		\caption{MOFA, IDP, IDR, and IDF$_1$ Results of Data Fusion for the Ten Clips from Table \ref{table:dataset}. (Unit: \%)}
		\begin{tabular}{c|l|cccc}
			\hline
			Video                    & Methods         & MOFA $\uparrow$ & IDP $\uparrow$ & IDR $\uparrow$ & IDF$_1 \uparrow$ \\ \hline \hline
			\multirow{5}{*}{clip-01} & EDDF            & 69.27           & 88.67          & 82.57          & 84.31            \\
			& MSDF \cite{liu2022intelligent}           & 68.35           & 88.61          & 82.11          & 83.84            \\
			& MMDST \cite{huang2021identity}          & 94.95           & \textbf{100.00}         & 95.41          & 97.42            \\
			& DeepSORVF (w/o) & 95.41           & \textbf{100.00}         & 95.41          & 97.65            \\
			& DeepSORVF       & \textbf{99.54}           & \textbf{100.00}         & \textbf{99.54}          & \textbf{99.77}            \\ \hline
			\multirow{5}{*}{clip-02} & EDDF            & 69.68           & 89.41          & 80.85          & 84.68            \\
			& MSDF \cite{liu2022intelligent}           & 69.68           & 89.41          & 80.85          & 84.68            \\
			& MMDST \cite{huang2021identity}          & 84.04           & \textbf{100.00}         & 87.77          & 92.18            \\
			& DeepSORVF (w/o) & 87.77           & \textbf{100.00}         & 87.77          & 93.48            \\
			& DeepSORVF       & \textbf{99.47}           & \textbf{100.00}         & \textbf{99.47}          & \textbf{99.73}            \\ \hline
			\multirow{5}{*}{clip-03} & EDDF            & 79.29           & 96.09          & 87.28          & 89.39            \\
			& MSDF \cite{liu2022intelligent}           & 81.66           & 96.09          & 87.28          & 90.49            \\
			& MMDST \cite{huang2021identity}          & 92.60           & \textbf{100.00}         & 92.60          & 96.16            \\
			& DeepSORVF (w/o) & 90.53           & \textbf{100.00}         & 90.53          & 95.03            \\
			& DeepSORVF       & \textbf{94.67}           & \textbf{100.00}         & \textbf{94.67}          & \textbf{97.26}            \\ \hline
			\multirow{5}{*}{clip-04} & EDDF            & 54.82           & 90.93          & 70.39          & 75.89            \\
			& MSDF \cite{liu2022intelligent}           & 53.73           & 90.66          & 72.37          & 75.95            \\
			& MMDST \cite{huang2021identity}          & 82.24           & 98.99          & 86.18          & 90.66            \\
			& DeepSORVF (w/o) & 85.53           & 99.00          & 86.40          & 92.27            \\
			& DeepSORVF       & \textbf{97.59}           & \textbf{99.33}          & \textbf{98.25}          & \textbf{98.79}            \\ \hline
			\multirow{5}{*}{clip-05} & EDDF            & 60.34           & 91.71          & 80.17          & 81.39            \\
			& MSDF \cite{liu2022intelligent}           & 57.78           & 91.13          & 78.89          & 80.09            \\
			& MMDST \cite{huang2021identity}          & 79.74           & \textbf{99.53}          & 89.55          & 91.01            \\
			& DeepSORVF (w/o) & 86.78           & 98.34          & 88.27          & 93.03            \\
			& DeepSORVF       & \textbf{92.96}           & 98.44          & \textbf{94.46}          & \textbf{96.41}            \\ \hline
			\multirow{5}{*}{clip-06} & EDDF            & 69.19           & 98.67          & 82.91          & 85.06            \\
			& MSDF \cite{liu2022intelligent}           & 71.15           & 98.66          & 82.63          & 85.88            \\
			& MMDST \cite{huang2021identity}          & 82.35           & \textbf{99.37}          & 87.96          & 91.41            \\
			& DeepSORVF (w/o) & 83.47           & 98.69          & 84.59          & 91.10            \\
			& DeepSORVF       & \textbf{92.16}           & 98.81          & \textbf{93.28}          & \textbf{95.97}            \\ \hline
			\multirow{5}{*}{clip-07} & EDDF            & 76.92           & 95.44          & 88.51          & 89.08            \\
			& MSDF \cite{liu2022intelligent}           & 75.74           & 95.21          & 87.92          & 88.48            \\
			& MMDST \cite{huang2021identity}          & 85.85           & \textbf{98.74}          & 92.73          & 93.37            \\
			& DeepSORVF (w/o) & 90.18           & 98.42          & 91.65          & 94.91            \\
			& DeepSORVF       & \textbf{93.03}           & 98.46          & \textbf{94.50}          & \textbf{96.44}            \\ \hline
			\multirow{5}{*}{clip-08} & EDDF            & 86.63           & 95.94          & 93.41          & 94.21            \\
			& MSDF \cite{liu2022intelligent}           & 87.38           & 95.57          & 93.41          & 94.21            \\
			& MMDST \cite{huang2021identity}          & 96.61           & \textbf{100.00}         & 97.55          & 98.48            \\
			& DeepSORVF (w/o) & 97.18           & \textbf{100.00}         & 97.18          & 98.57            \\
			& DeepSORVF       & \textbf{98.49}           & \textbf{100.00}         & \textbf{98.49}          & \textbf{99.24}            \\ \hline
			\multirow{5}{*}{clip-09} & EDDF            & 79.08           & 96.42          & 87.91          & 89.61            \\
			& MSDF \cite{liu2022intelligent}           & 81.80           & 96.42          & 87.77          & 89.47            \\
			& MMDST \cite{huang2021identity}          & 92.66           & \textbf{99.15}          & 94.57          & 96.53            \\
			& DeepSORVF (w/o) & 92.93           & 98.86          & 94.02          & 96.38            \\
			& DeepSORVF       & \textbf{94.70}           & 98.34          & \textbf{96.33}          & \textbf{97.32}            \\ \hline
			\multirow{5}{*}{clip-10} & EDDF            & 87.56           & 98.05          & 89.33          & 93.49            \\
			& MSDF \cite{liu2022intelligent}           & 88.44           & 98.54          & 89.78          & 93.95            \\
			& MMDST \cite{huang2021identity}          & 88.44           & \textbf{100.00}         & 89.56          & 93.94            \\
			& DeepSORVF (w/o) & 89.56           & \textbf{100.00}         & 89.56          & 94.49            \\
			& DeepSORVF       & \textbf{97.78}           & \textbf{100.00}         & \textbf{97.78}          & \textbf{98.88}            \\ \hline
			\multirow{5}{*}{Average} & EDDF            & 73.28           & 94.13          & 84.33          & 86.71            \\
			& MSDF \cite{liu2022intelligent}           & 73.57           & 94.03          & 84.30          & 86.70            \\
			& MMDST \cite{huang2021identity}          & 87.95           & \textbf{99.60}          & 91.39          & 94.12            \\
			& DeepSORVF (w/o) & 89.93           & 99.33          & 90.54          & 94.69            \\
			& DeepSORVF       & \textbf{96.04}           & 99.34          & \textbf{96.68}          & \textbf{97.98}             \\ \hline
		\end{tabular}\label{table:fusion}
	\end{table}
	\begin{figure*}[h]
		\centering
		\includegraphics[width=0.97\linewidth]{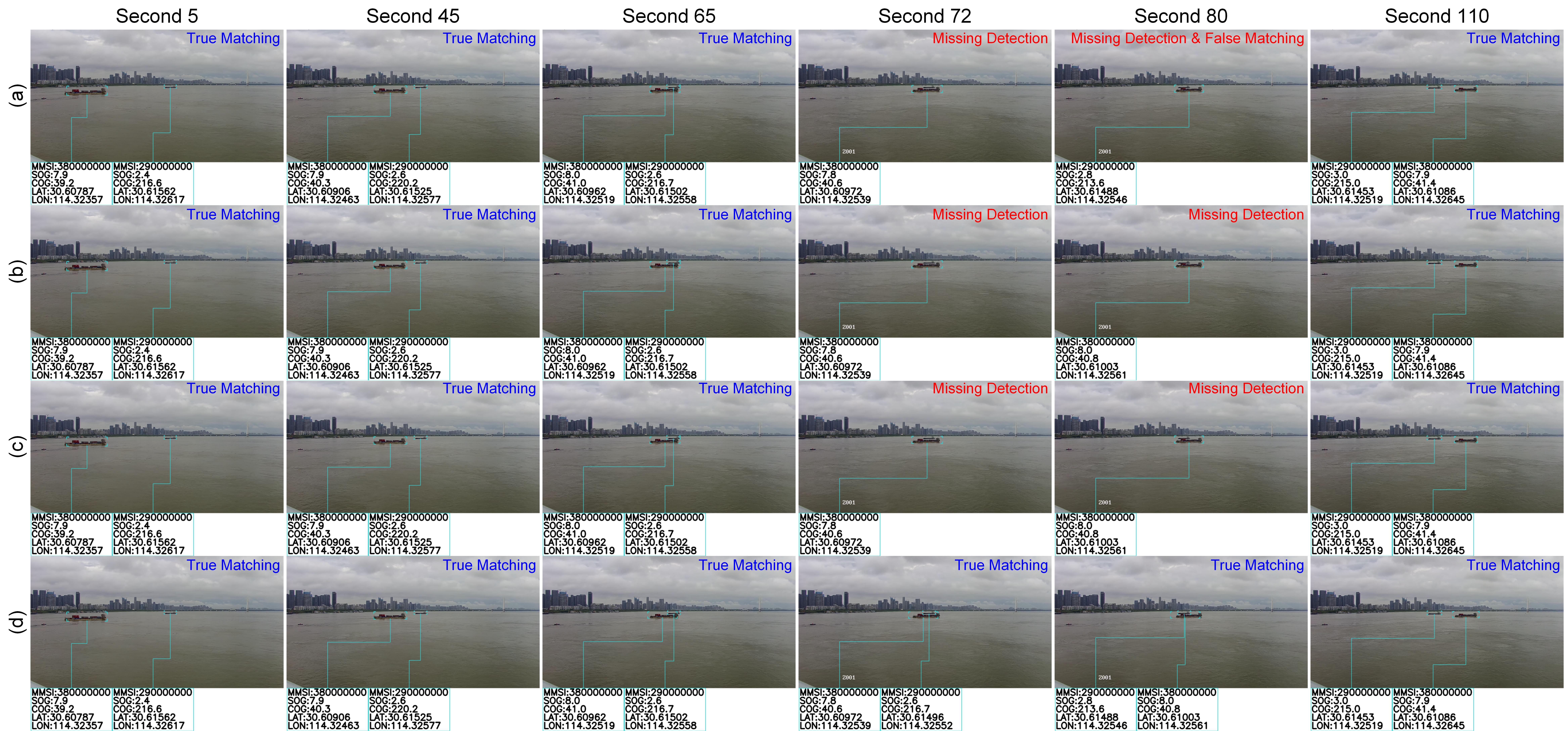}
		\caption{Visual comparisons of fusion results on the dataset captured by the bridge region camera from Table \ref{table:dataset}. From top to bottom: visual fusion results generated by (a) MSDF \cite{liu2022intelligent}, (b) MMDST \cite{huang2021identity}, (c) DeepSORVF without the anti-occlusion strategy, and (d) DeepSORVF, respectively.}
		\label{fig:Bridge}
	\end{figure*}
	\begin{figure*}[ht]
		\centering
		\includegraphics[width=0.97\linewidth]{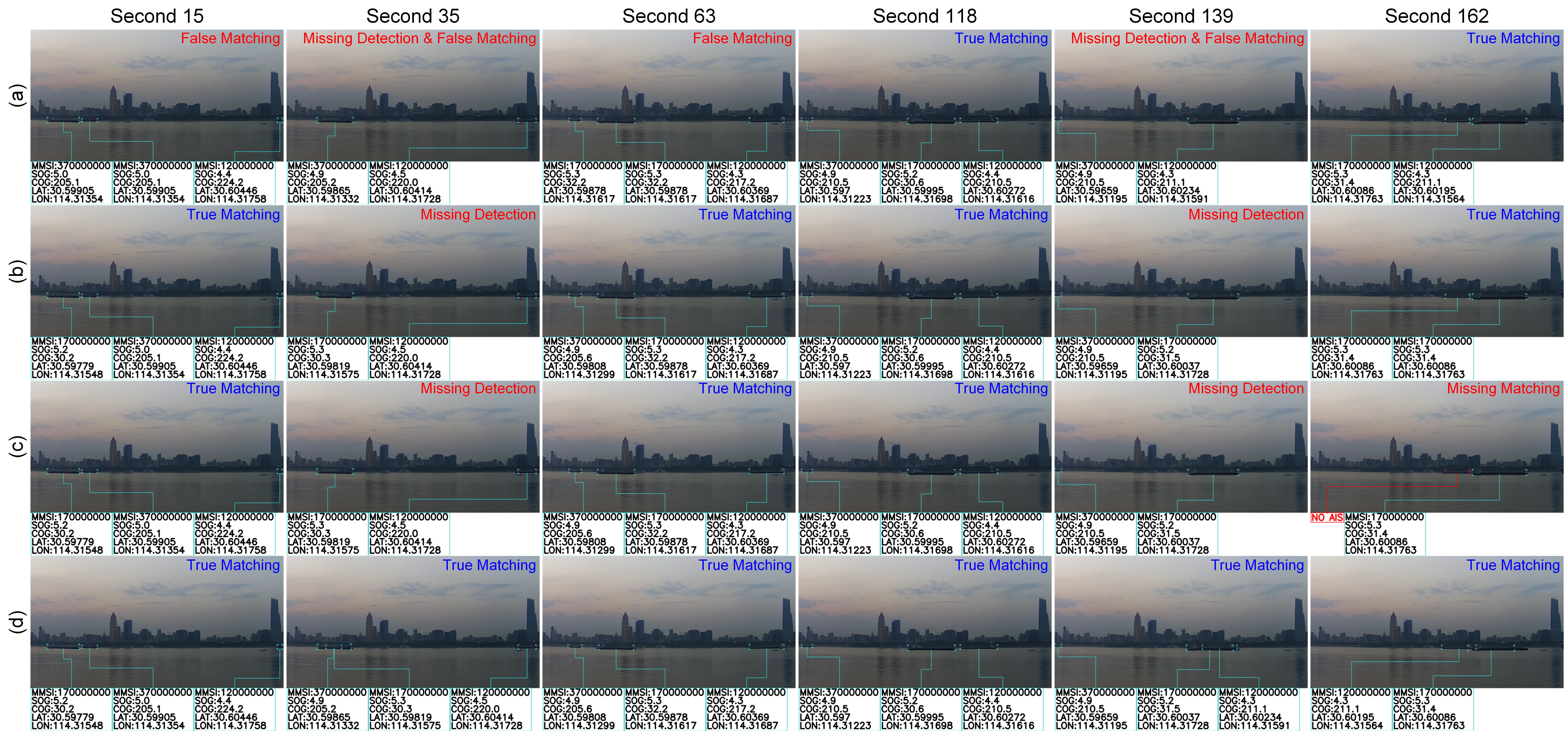}
		\caption{Visual comparisons of fusion results on the dataset captured by the riverside camera from Table \ref{table:dataset}. From top to bottom: visual fusion results generated by (a) MSDF \cite{liu2022intelligent}, (b) MMDST \cite{huang2021identity}, (c) DeepSORVF without the anti-occlusion strategy, and (d) DeepSORVF, respectively.}
		\label{fig:Riverside1}
	\end{figure*}
	\subsection{Experiments on Data Fusion}
	In this section, we implement the data fusion experiment to compare various methods, i.e., Euclidean distance-based data fusion (EDDF),  multi-source data fusion (MSDF) \cite{liu2022intelligent}, multi-modal data-based ship tracking (MMDST) \cite{huang2021identity}, DeepSORVF (w/o) without the anti-occlusion strategy, and DeepSORVF. In particular, the EDDF calculates the Euclidean distance between the pixel position points at the current moment for similarity measurement and employs the near-matching mechanism. For the point matching-based MSDF and MMDST, we only replace our asynchronous vessel trajectory matching part with its corresponding matching method to compare the fusion effect under the premise of consistent detectors. Furthermore, all methods only process data once per second.

	\subsubsection{Evaluation Metric}
	To evaluate the performance of data fusion, we first use a variant of multi-object tracking accuracy ($\mathrm{MOTA}$) \cite{bernardin2008evaluating} as the evaluation metric and name it multi-object fusion accuracy ($\mathrm{MOFA}$), i.e.,
	\begin{equation}\label{eq:MOFA}
	\mathrm{MOFA} = 1-\frac{FN_{\mathrm{mmsi}}+FP_{\mathrm{mmsi}}}{GT_{\mathrm{mmsi}}},
	\end{equation}
	where $\mathrm{mmsi}$ represents the identity of the vessel of interest (MMSI), $FP_\mathrm{mmsi}$, $FN_\mathrm{mmsi}$, and $GT_\mathrm{mmsi}$ are the number of the MMSI false positive, MMSI false negative, and MMSI ground truth, respectively. Furthermore, the identification precision ($\mathrm{IDP}$), identification recall ($\mathrm{IDR}$), and identification F1 ($\mathrm{IDF}_1$) are also employed as evaluation metrics. The $\mathrm{IDP}$, $\mathrm{IDR}$, and $\mathrm{IDF}_1$ can be given by
	\begin{equation}\label{eq:IDP}
	\mathrm{IDP} = \frac{TP_{\mathrm{id}}}{TP_{\mathrm{id}}+FP_{\mathrm{id}}},
	\end{equation}
	\begin{equation}\label{eq:IDR}
	\mathrm{IDR} = \frac{TP_{\mathrm{id}}}{TP_{\mathrm{id}}+FN_{\mathrm{id}}},
	\end{equation}
	\begin{equation}\label{eq:IDF1}
	\mathrm{IDF}_1 = \frac{2TP_{\mathrm{id}}}{2TP_{\mathrm{id}}+FP_{\mathrm{id}}+FN_{\mathrm{id}}},
	\end{equation}
	where $TP_{\mathrm{id}}$, $FP_{\mathrm{id}}$, and $FN_{\mathrm{id}}$ are the numbers of the ID true positive, ID false positive, and ID false negative, respectively. In particular, the $\mathrm{id}$ is replaced with the identity of the vessel of interest (MMSI) in the data fusion evaluation. Please refer to  Refs. \cite{bernardin2008evaluating, ristani2016performance} for more details on the $\mathrm{MOTA}$, $\mathrm{IDP}$, $\mathrm{IDR}$, and $\mathrm{IDF}_1$. Generally, higher $\mathrm{MOFA}$, $\mathrm{IDP}$, $\mathrm{IDR}$, and $\mathrm{IDF}_1$ mean better fusion performance.
	\begin{figure*}[t]
		\centering
		\includegraphics[width=1\linewidth]{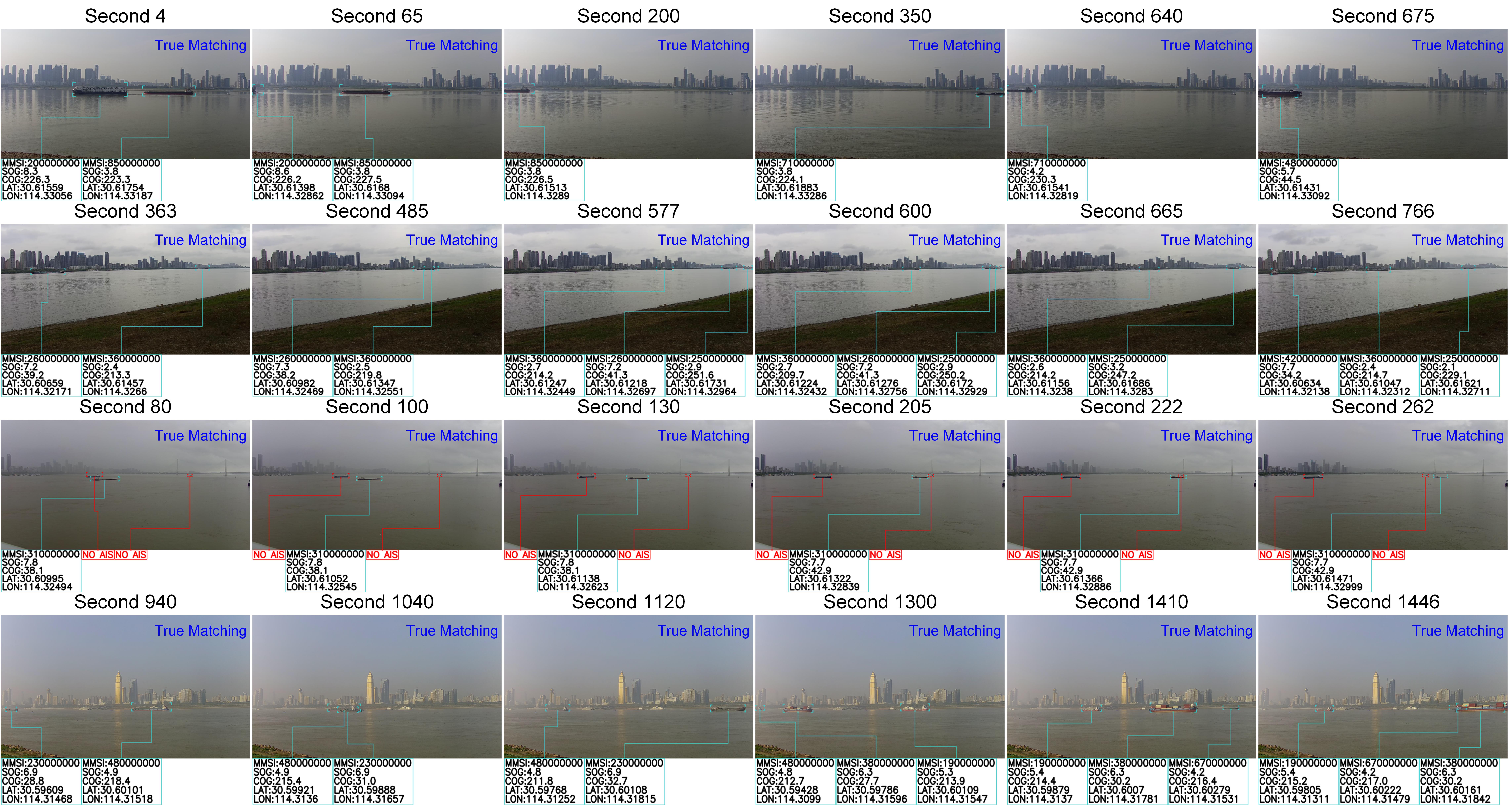}
		\caption{Visual fusion results of our DeepSORVF on the FVessel dataset from Table \ref{table:FVessel}.}
		\label{fig:fusion}
	\end{figure*}
	\subsubsection{Fusion Results on Ten Clips}
	Table \ref{table:fusion} displays the evaluation results on all clips. It can be found that EDDF and MSDF perform poorly. Especially for clip-04, the MOFA is only 54.82\% for EDDF and 53.73\% for MSDF. The poor fusion effect stems from the fact that these methods only consider the current information without associating the historical feature. By considering the displacement direction of AIS- and video-based vessel trajectories, the MMDST greatly improves the fusion effect. However, the two DeepSORVFs based on the vessel motion trajectory matching have better performance by comparison. Particularly after implementing the anti-occlusion strategy, the performance of our DeepSORVF has improved considerably across all metrics.
	\setlength{\tabcolsep}{12pt}
	\begin{table}[t]
		\scriptsize
		\centering
		\caption{MOFA, IDP, IDR, and IDF$_1$ Results of Data Fusion for the FVessel from Table \ref{table:FVessel}. (Unit: \%)}
		\begin{tabular}{c|cccc}
			\hline
			Video    & MOFA $\uparrow$ & IDP $\uparrow$ & IDR $\uparrow$ & IDF$_1 \uparrow$ \\ \hline \hline
			video-01 & 79.94           & 89.35          & 90.76          & 90.05            \\
			video-02 & 73.19           & 83.27          & 91.60          & 87.23            \\
			video-03 & 96.45           & 99.23          & 97.20          & 98.20            \\
			video-04 & 98.08           & 99.45          & 98.63          & 99.03            \\
			video-05 & 89.19           & 93.46          & 95.91          & 94.67            \\
			video-06 & 91.17           & 96.04          & 95.08          & 95.56            \\
			video-07 & 96.81           & 99.59          & 97.21          & 98.39            \\
			video-08 & 82.28           & 99.64          & 82.58          & 90.31            \\
			video-09 & 98.45           & 100.00         & 98.45          & 99.22            \\
			video-10 & 88.74           & 90.42          & 99.26          & 94.63            \\
			video-11 & 97.66           & 99.29          & 98.36          & 98.83            \\
			video-12 & 95.45           & 99.06          & 96.36          & 97.69            \\
			video-13 & 84.82           & 94.82          & 89.72          & 92.20            \\
			video-14 & 93.10           & 97.82          & 95.22          & 96.50            \\
			video-15 & 95.88           & 97.19          & 98.74          & 97.96            \\
			video-16 & 98.68           & 100.00         & 98.68          & 99.33            \\
			video-17 & 90.02           & 93.80          & 96.39          & 95.08            \\
			video-18 & 74.49           & 83.57          & 92.72          & 87.91            \\
			video-19 & 96.62           & 98.31          & 98.31          & 98.31            \\
			video-20 & 96.74           & 98.66          & 98.07          & 98.36            \\
			video-21 & 76.43           & 87.03          & 89.82          & 88.40            \\
			video-22 & 96.82           & 99.35          & 97.45          & 98.39            \\
			video-23 & 94.71           & 98.91          & 95.77          & 97.31            \\
			video-24 & 94.70           & 98.34          & 96.33          & 97.32            \\
			video-25 & 91.49           & 97.66          & 93.73          & 95.66            \\
			video-26 & 97.44           & 99.11          & 98.32          & 98.72            \\ \hline
			Average  & 91.13           & 95.90          & 95.41          & 95.59            \\ \hline
		\end{tabular}\label{table:fusion_FVessel}
	\end{table}
	To provide a more understandable explanation, we display two examples of data fusion obtained by MSDF, MMDST, DeepSORVF (w/o), and DeepSORVF shown in Figs. \ref{fig:Bridge} and \ref{fig:Riverside1}. Specifically, Fig. \ref{fig:Bridge} displays the visualized data fusion result captured by the bridge region camera. Since the MSDF only considers the vessel characteristic at the current moment, the vessel information is more likely to be matched incorrectly. In the 80-th second, the MSDF, MMDST, and DeepSORVF (w/o) are unable to match the vessel identification information since the detector fails to identify the partially occluded target. For the data collected by the riverside, the visual vessels are often more severely occluded, resulting in the complete disappearance of target features. By analyzing Fig. \ref{fig:Riverside1}, the MSDF, MMDST, and DeepSORVF (w/o) will produce more missing detection and false matching. It is worth mentioning that the vessel occlusion will also affect the trajectory feature extraction and cause the feature matching failure. In contrast, the proposed DeepSORVF with the anti-occlusion strategy has a more stable data fusion effect and is suitable for a variety of scenarios.
	\subsubsection{Fusion Results on FVessel Dataset}
	Our DeepSORVF is also used to process more data in the FVessel dataset and calculate the $\mathrm{MOFA}$, $\mathrm{IDP}$, $\mathrm{IDR}$, and $\mathrm{IDF}_1$. Table \ref{table:fusion_FVessel} and Fig. \ref{fig:fusion} display the metric calculation results and the visualized fusion results, respectively. It can be found that the proposed method has stable fusion performance. The fusion accuracy (MOFA) of our method is between 73.19\% and 98.68\% and the average is 91.13\%. In the evaluation of the other three metrics, our method also has a good performance. Through the comparison in Fig. \ref{fig:fusion}, the results generated by our DeepSORVF are accurate and stable. The superiority of the proposed method benefits from the accurate prediction of the vessel bounding box by the anti-occlusion tracking method under the occlusion condition and the accurate matching based on the trajectory series.
	\setlength{\tabcolsep}{12pt}
	\begin{table}[t]
		\scriptsize
		\centering
		\caption{Precision and Recall Results of Vessel Detection for the Ten Clips from Table \ref{table:dataset}. (Unit: \%)}
		\begin{tabular}{l|cccccc}
			\hline
			Detector                                             & Data Fusion    & Precision $\uparrow$ & Recall $\uparrow$ \\ \hline \hline
			\multirow{2}{*}{Faster-RCNN   \cite{ren2016faster}}  & \XSolidBrush   & 89.04                & 90.74             \\
			& \CheckmarkBold & \textbf{90.03}                & \textbf{96.48}             \\ \hline
			\multirow{2}{*}{SSD   \cite{liu2016ssd}}             & \XSolidBrush   & 94.98                & 88.14             \\
			& \CheckmarkBold & \textbf{95.02}                & \textbf{95.62}             \\ \hline
			\multirow{2}{*}{YOLOv4 \cite{bochkovskiy2020yolov4}} & \XSolidBrush   & 96.43                & 76.97             \\
			& \CheckmarkBold & \textbf{96.74}                & \textbf{84.85}             \\ \hline
			\multirow{2}{*}{YOLOv5 \cite{jocher2021ultralytics}} & \XSolidBrush   & 98.78                & 83.70             \\
			& \CheckmarkBold & \textbf{99.47}                & \textbf{91.29}             \\ \hline
			\multirow{2}{*}{YOLOX \cite{ge2021yolox}}            & \XSolidBrush   & 98.91                & 94.13             \\
			& \CheckmarkBold & \textbf{99.20}                & \textbf{99.40}               \\ \hline
		\end{tabular}\label{table:detection}
	\end{table}
	\subsection{Influence of Data Fusion on Vessel Detection and Tracking}
	In our proposed method, the result of trajectory matching is fed as prior knowledge to the vessel detection and tracking tasks at the next moment for promoting the video-based vessel trajectory extraction. To verify that our proposed data fusion method can improve vessel detection and tracking performance, we conduct massive experiments on ten clips. In particular, we select five different deep neural networks as detectors, i.e., Faster-RCNN \cite{ren2016faster}, SSD \cite{liu2016ssd}, YOLOv4 \cite{bochkovskiy2020yolov4}, YOLOv5 \cite{jocher2021ultralytics}, and YOLOX \cite{ge2021yolox}. Each detector has two versions, i.e., ``Detection'' and ``Detection + Data Fusion''. Furthermore, all methods only process data once per second.
	\subsubsection{Evaluation Metric}
	To evaluate the performance of vessel detection, we select the $\mathrm{Precision}$ and $\mathrm{Recall}$ as evaluation metrics. Let $TP$, $FP$, and $FN$ denote the number of the true positive, false positive, and false negative, the $\mathrm{Precision}$ and $\mathrm{Recall}$ can be given by
	\begin{equation}\label{eq:Precision}
		\mathrm{Precision} = \frac{TP}{TP+FP},
	\end{equation}
	\begin{equation}\label{eq:Recall}
		\mathrm{Recall} = \frac{TP}{TP+FN}.
	\end{equation}

	For vessel tracking, we tend to use MOTA as an evaluation metric, which can be defined as
	\begin{equation}\label{eq:MOTA}
	\mathrm{MOTA} = 1-\frac{FP+FN+ID_s}{GT},
	\end{equation}
	where $FP$, $FN$, $ID_s$, and $GT$ represent the numbers of the false positive, false negative, $ID$ switch, and ground truth, respectively. Furthermore, we also adopt the $\mathrm{IDP}$, $\mathrm{IDR}$, and $\mathrm{IDF}_1$ metrics. Theoretically, better detection results have higher $\mathrm{Precision}$ and $\mathrm{Recall}$, and better tracking results have higher $\mathrm{MOTA}$, $\mathrm{IDP}$, $\mathrm{IDR}$, and $\mathrm{IDF}_1$.

	\subsubsection{Vessel Detection and Tracking on Ten Clips}
	Table \ref{table:detection} compares the detection $\mathrm{Precision}$ and $\mathrm{Recall}$ of various detectors on ten clips. Due to the mutual occlusion between the targets, some vessel characteristics are easily hidden by another vessel. Therefore, detectors often suffer from missing detection, resulting in higher $FN$ and poorer $\mathrm{Recall}$. In most cases, detectors are prone to produce false detection boxes in vessel encounter regions due to the overlapping of multiple vessel features. These false detection boxes will produce higher $FP$ and poorer $\mathrm{Precision}$. In contrast, the proposed anti-occlusion method based on data fusion results can improve the performance of various detectors.
	\setlength{\tabcolsep}{6pt}
	\begin{table}[t]
		\scriptsize
		\centering
		\caption{MOTA, IDP, IDR, and IDF$_1$ Results of Vessel Tracking for the Ten Clips from Table \ref{table:dataset}. (Unit: \%)}
		\begin{tabular}{l|ccccc}
			\hline
			Detector                                             & Data Fusion    & MOTA $\uparrow$ & IDP $\uparrow$ & IDR $\uparrow$ & IDF$_1 \uparrow$ \\ \hline \hline
			\multirow{2}{*}{Faster-RCNN   \cite{ren2016faster}}  & \XSolidBrush   & 75.19           & 68.03          & 69.00          & 68.43            \\
			& \CheckmarkBold & \textbf{84.43}           & \textbf{81.81}          & \textbf{87.95}          & \textbf{84.58}            \\ \hline
			\multirow{2}{*}{SSD   \cite{liu2016ssd}}             & \XSolidBrush   & 81.36           & 84.49          & 78.77          & 81.14            \\
			& \CheckmarkBold & \textbf{89.69}           & \textbf{90.03}          & \textbf{91.03}          & \textbf{90.13}            \\ \hline
			\multirow{2}{*}{YOLOv4 \cite{bochkovskiy2020yolov4}} & \XSolidBrush   & 70.92           & 71.61          & 56.66          & 63.08            \\
			& \CheckmarkBold & \textbf{80.50}           & \textbf{80.44}          & \textbf{70.18}          & \textbf{74.77}            \\ \hline
			\multirow{2}{*}{YOLOv5 \cite{jocher2021ultralytics}} & \XSolidBrush   & 81.72           & 88.68          & 75.08          & 81.12            \\
			& \CheckmarkBold & \textbf{90.40}           & \textbf{94.48}          & \textbf{86.50}          & \textbf{90.05}            \\ \hline
			\multirow{2}{*}{YOLOX \cite{ge2021yolox}}            & \XSolidBrush   & 92.61           & 89.80          & 85.36          & 87.50            \\
			& \CheckmarkBold & \textbf{98.61}           & \textbf{97.72}          & \textbf{97.92}          & \textbf{97.82}            \\ \hline
		\end{tabular}\label{table:tracking}
	\end{table}
	To compare the tracking performance, Table \ref{table:tracking} further shows the $\mathrm{MOTA}$, $\mathrm{IDP}$, $\mathrm{IDR}$, and $\mathrm{IDF}_1$ results of various detectors on ten clips. In contrast, the proposed data fusion method can significantly improve the tracking performance of all five detectors and reduce the number of missing and false detection. The performance improvement benefits from the proposed anti-occlusion method based on data fusion results. The proposed method can achieve more stable vessel tracking during the occlusion.
	\begin{figure}[t]
		\centering
		\includegraphics[width=1\linewidth]{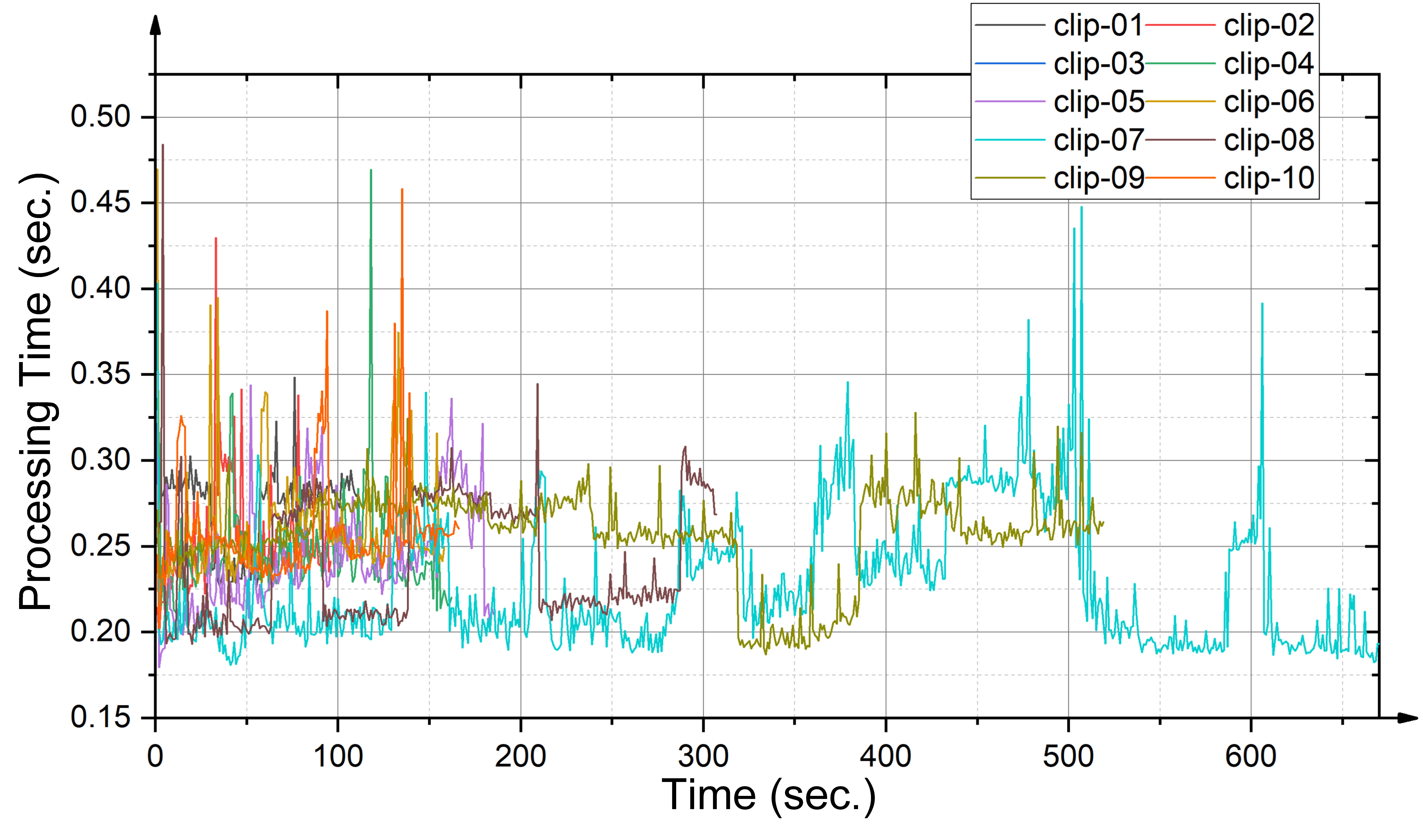}
		\caption{Processing time of one-second data on the ten clips from Table \ref{table:dataset}.}
		\label{fig:time}
	\end{figure}
	\subsection{Running Time Analysis}
	\setlength{\tabcolsep}{14pt}
	\begin{table}[t]
		\scriptsize
		\centering
		\caption{Processing Time of One-Second Data (Mean $\pm$ Std) on the Ten Clips from Table \ref{table:dataset}. (Unit: Sec.)}
		\begin{tabular}{c|cc}
			\hline
			Video   & Video Length & Processing Time   \\ \hline \hline
			clip-01 & 01m51s       & 0.2721$\pm$0.0230 \\
			clip-02 & 01m36s       & 0.2526$\pm$0.0323 \\
			clip-03 & 03m42s       & 0.2998$\pm$0.0413 \\
			clip-04 & 02m42s       & 0.2462$\pm$0.0289 \\
			clip-05 & 03m05s       & 0.2444$\pm$0.0321 \\
			clip-06 & 02m38s       & 0.2642$\pm$0.0344 \\
			clip-07 & 11m10s       & 0.2279$\pm$0.0391 \\
			clip-08 & 05m07s       & 0.2404$\pm$0.0364 \\
			clip-09 & 08m39s       & 0.2569$\pm$0.0256 \\
			clip-10 & 02m46s       & 0.2579$\pm$0.0304 \\ \hline
			Average &   --         & 0.2562            \\ \hline
		\end{tabular}\label{table:time}
	\end{table}
	The time complexity of the proposed method is a critical metric, which directly determines whether it can be used in actual engineering. In this work, we only process the data once per second to ensure practicability. Therefore, we are unable to use the frame per second (FPS) as an evaluation metric. Meanwhile, since the proposed method considers trajectory features, the time complexity is also related to the number and length of AIS- and video-based vessel trajectories. Consequently, it is also inaccurate to calculate the running time of a single image. Finally, we compute the processing time of one-second data for ten clips in Table \ref{table:dataset}. The processing time of our method for each clip is shown in Fig. \ref{fig:time} and Table \ref{table:time}. It can be seen that our DeepSORVF has low time complexity and high practicability. It can process one second of data in 0.175-0.500 seconds and 0.2562 seconds on average.
	\subsection{Discussion}
	Although our proposed method adopts the prior knowledge-driven anti-occlusion tracking method and trajectory matching method to effectively improve the accuracy of data fusion, our method still has some limitations. In this section, we use the multiple object fusion accuracy (MOFA) and multiple object fusion precision (MOFP) as evaluation metrics, where the MOFP is a variant of the Multiple Object Tracking Precision (MOTP) \cite{bernardin2008evaluating} in the data fusion task. The MOFP can be given by
	\begin{equation}\label{eq:MOTP}
	\mathrm{MOFP} = \frac{\varSigma_{t,i}\mathcal{D}^{t,i}_{\mathrm{mmsi}}}{\varSigma_{t}N^{t}_{\mathrm{mmsi}}},
	\end{equation}
	where $\mathcal{D}^{t,i}_{\mathrm{mmsi}}$ denotes the distance of the $i$-th MMSI matching pair in the $t$-th second, $N^{t}_{\mathrm{mmsi}}$ is the number of matches in the $t$-th second. Theoretically, a better fusion effect has higher MOFA and lower MOFP.
	\begin{figure}[t]
		\centering
		\includegraphics[width=0.98\linewidth]{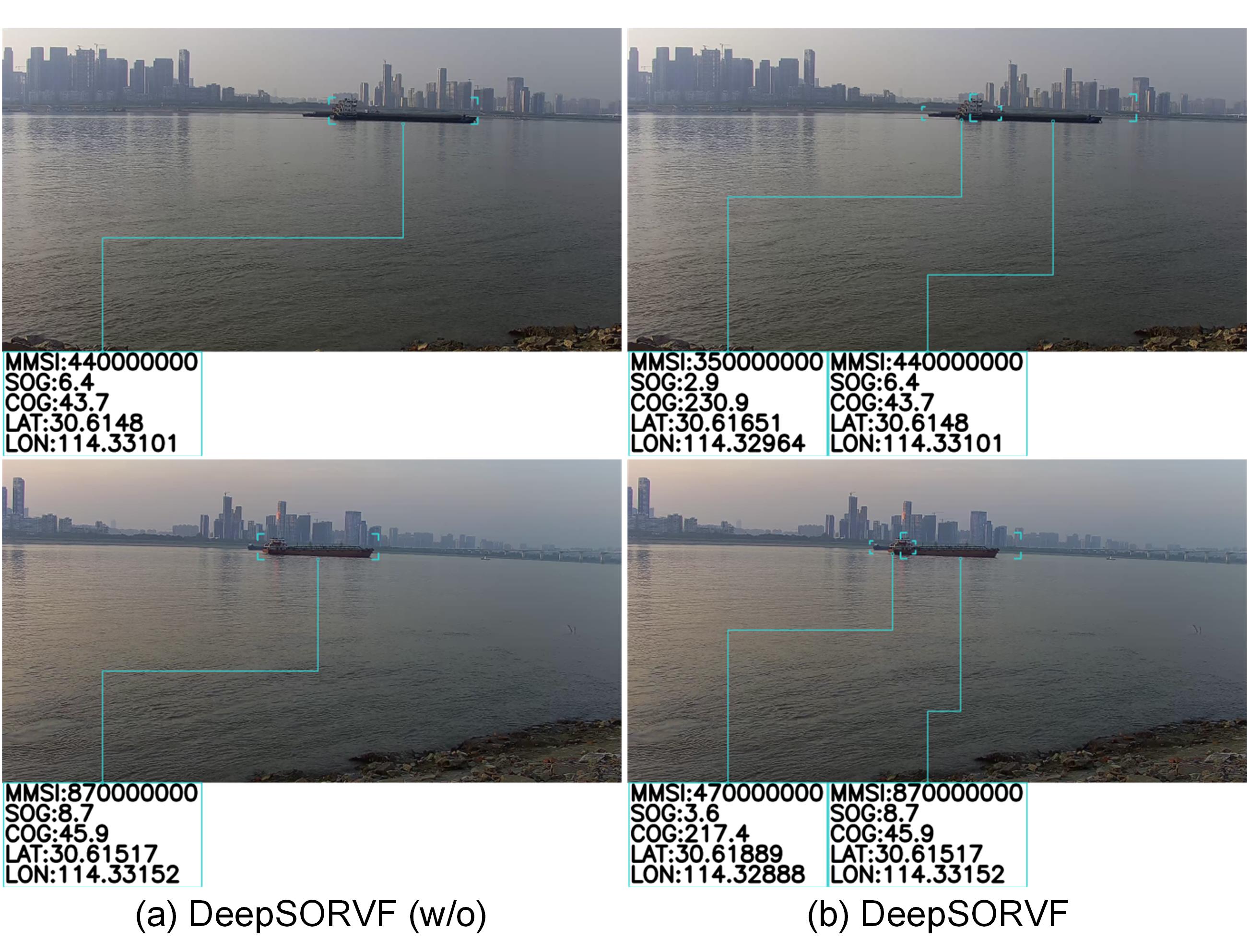}
		\caption{Visual comparisons of fusion results on the dataset from Table \ref{table:dataset}. DeepSORVF (w/o) represents our DeepSORVF without the anti-occlusion strategy.}
		\label{fig:Limitation}
	\end{figure}
	\setlength{\tabcolsep}{12pt}
	\begin{table}[t]
		\scriptsize
		\centering
		\caption{MOFA (\%) and MOFP Results of Data Fusion for the clip-02 and clip-03 from Table \ref{table:dataset}. DeepSORVF (w/o) Represents Our DeepSORVF Without the Anti-Occlusion Strategy.}
		\begin{tabular}{c|l|cc}
			\hline
			Video                    & Methods         & MOFA $\uparrow$ & MOFP $\downarrow$ \\ \hline \hline
			\multirow{2}{*}{clip-02} & DeepSORVF (w/o) & 87.77           & \textbf{0.110}             \\
			& DeepSORVF       & \textbf{99.47}           & 0.140             \\ \hline
			\multirow{2}{*}{clip-03} & DeepSORVF (w/o) & 90.53           & \textbf{0.241}             \\
			& DeepSORVF       & \textbf{94.67}           & 0.252             \\ \hline
			\multirow{2}{*}{Average} & DeepSORVF (w/o) & 88.92           & \textbf{0.175}             \\
			& DeepSORVF       & \textbf{97.02}           & 0.196              \\ \hline
		\end{tabular}\label{table:limitation}
	\end{table}
	Using clip-02 and clip-03 as examples, we compute their MOFA and MOFP. As shown in Table \ref{table:limitation}, it can be found that the proposed anti-occlusion tracking method can significantly improve the accuracy of data fusion by comparing the MOFA. However, the DeepSORVF is slightly inferior to the DeepSORVF (w/o) without the anti-occlusion strategy in the bounding box localization precision evaluated by the MOFP. The more intuitive comparisons before and after the use of the anti-occlusion strategy are illustrated in Fig. \ref{fig:Limitation}. Our DeepSORVF with the anti-occlusion strategy can predict the vessel position and accurately match the occluded vessel information. However, the predicted bounding boxes still have some degree of bias in complex occlusion conditions. This deviation is mostly attributable to the inaccurate estimation of AIS and visual motion characteristics. When a vessel travels away from the camera, for instance, its visual movement speed generally slows and the object gets smaller. To further improve the vessel anti-occlusion performance, our future work will take into account the changing features of the moving vessels in the visual data.
	\section{Conclusion}
	\label{con}
	In this paper, we proposed a deep learning-based simple online and real-time vessel data fusion method (named DeepSORVF). The DeepSORVF could pair the vessel features of AIS with visual targets. Due to the fact that reciprocal occlusion between vessel targets may readily interfere with video-based trajectory extraction, we suggested a prior knowledge-driven anti-occlusion tracking method. Meanwhile, a novel asynchronous trajectory matching method was designed for robust data fusion. Comprehensive experiments on vessel detection, vessel tracking, data fusion, and running time analysis have demonstrated the superior performance of our DeepSORVF on the newly-developed FVessel dataset.

\bibliographystyle{IEEEtran}
\bibliography{tits.bib}


\end{document}